\documentclass[10pt,twocolumn,letterpaper]{article}

\usepackage[pagenumbers]{cvpr}      
\usepackage{multicol}
\usepackage{multirow}
\usepackage{algorithm}
\usepackage{algorithmic}
\usepackage{parskip}
\usepackage{booktabs, graphicx} 
\usepackage[table]{xcolor}
\usepackage{subcaption}

\usepackage[most]{tcolorbox}
\usepackage{enumitem}

\definecolor{cvprblue}{rgb}{0.21,0.49,0.74}
\usepackage[pagebackref,breaklinks,colorlinks,allcolors=cvprblue]{hyperref}

\def\confName{CVPR}
\def\confYear{2027}

\title{Beyond Random Starts: Amortized Initialization for \\ Adversarial Attacks on Vision-Language Models}

\author{
Haz Sameen Shahgir$^{1}$,
Ali Asif Khan$^{2}$,
K. M. Mehemud Azad$^{2}$,
Dongwei Lyu$^{3}$,
Zhipeng Wei$^{3}$ \\
Rifat Shahriyar$^{2}$, 
Yufei Li$^{4}$,
Xian Chen$^{4}$,
Xiaohan Wei$^{4}$,
Yunchen Pu$^{4}$,
Fei Tian$^{4}$,
Chonglin Sun$^{4}$ \\
Frank Shyu$^{4}$,
Sandeep Pandey$^{4}$,
Luke Simon$^{4}$,
Xi Liu$^{4}$,
N. Benjamin Erichson$^{5,6*}$,
Yue Dong$^{1}$\thanks{Corresponding authors: N. Benjamin Erichson (\url{erichson@lbl.gov}) and Yue Dong (\url{yue.dong@ucr.edu}).} \\[0.5em]
$^{1}$UC Riverside,
$^{2}$BUET,
$^{3}$UC Berkeley,
$^{4}$Meta \\
$^{5}$International Computer Science Institute,
$^{6}$Lawrence Berkeley National Laboratory
}

\begin{document}
\maketitle
\begin{abstract}

Gradient-based transferable adversarial attacks achieve state-of-the-art attack success rates against black-box vision-language models (VLMs). However, they require hundreds of costly optimization steps for each image pair. One possible reason is that these attacks typically start from random perturbations that contain no information about the source or target image. In this work, we study whether amortized adversarial generation can provide better initializations for gradient-based attacks. By learning across many image pairs, amortization allows a generator to produce adversarial perturbations for new inputs without optimizing each perturbation from scratch. We introduce SPARK, a lightweight looped perturbation generator conditioned on both the source and target images. SPARK repeatedly applies the same network to refine a perturbation, producing effective adversarial attacks in a few forward passes without requiring gradients through surrogate vision encoders at inference time. We show that these perturbations provide effective warm starts for existing gradient-based attacks, substantially improving attack success rates while reducing the number of optimization steps required. These improvements generalize across different attack methods, VLMs, and unseen image categories. SPARK also produces effective standalone attacks at a fraction of the computational cost of gradient-based optimization. \footnote{Code and model weights are available at \url{https://github.com/Patchwork53/SPARK}}

\end{abstract}    
\section{Introduction}

\begin{figure*}[!t]
    \centering
    \includegraphics[width=0.9\linewidth]{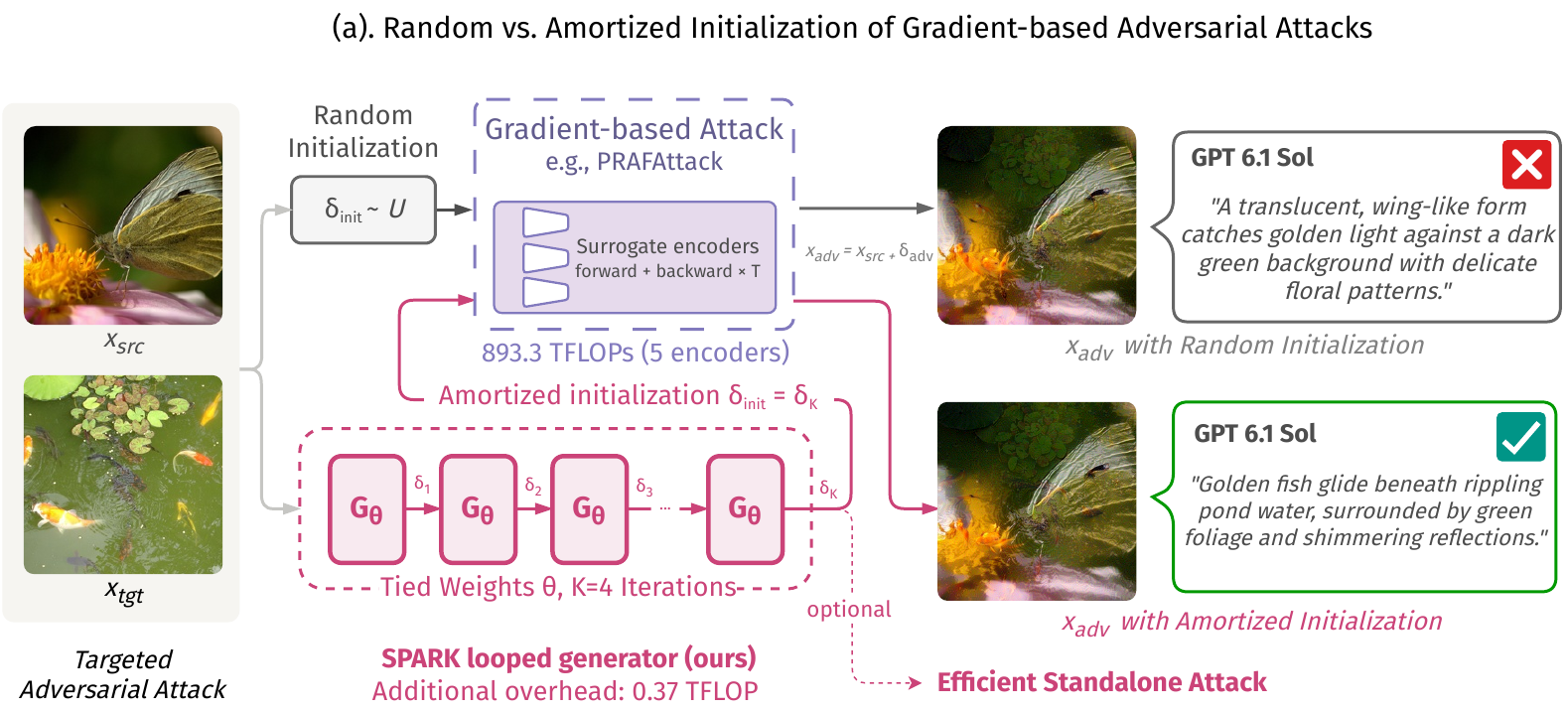}

    \caption{SPARK uses amortized adversarial generation to warm-start gradient-based attacks. Conventional attacks start from random perturbations and iteratively optimize them using gradients from surrogate vision encoders. In contrast, SPARK learns from source-target image pairs to generate adversarial perturbations through a few passes of a network with shared weights. These perturbations can be used as standalone attacks or to warm-start gradient-based optimization, improving attack success at little additional computational cost.}

    \label{fig:spark-overview}
\end{figure*}

Vision-language models (VLMs) enable AI agents to interpret visual content at scale, from extracting product information on websites to solving visual CAPTCHAs. While these capabilities support useful applications, they also enable automated scraping and can bypass mechanisms intended to restrict automated access. Adversarial image perturbations offer a potential defense by disrupting how VLMs interpret visual content, with applications including CAPTCHA hardening \citep{Kwon2020RobustCI}, facial privacy protection \citep{cherepanova2021lowkey}, and preventing unauthorized AI training \citep{huang2021unlearnable,shan2024nightshade,shan2023glaze}. Adversarial perturbations are also used to improve model robustness through adversarial training \citep{adversarialtraining}. In many of these applications, we do not know which models will process the images. The perturbations must therefore transfer across different VLMs, including proprietary models whose parameters and gradients are inaccessible \citep{attackvlm,dong2023robust}. Protecting large collections of images also requires generating adversarial perturbations at scale, making computational efficiency an important practical consideration.

State-of-the-art transferable attacks, such as MAttack-v2 \citep{mattackv2}, MPCAttack \citep{mpcattack}, and PRAFAttack \citep{prafattack}, rely on iterative gradient-based optimization. Given a source image and a target image, these methods optimize a small perturbation that makes the source image resemble the target in the representation spaces of multiple surrogate vision encoders. The perturbation is typically initialized randomly and updated using gradients from the surrogate encoders, while remaining within a prescribed $\ell_\infty$ budget. However, a random initialization contains no information about the source or target image, which may explain why these attacks require hundreds of optimization steps to achieve high attack success rates. Indeed, attack success rates only begin to exhibit diminishing returns around 300 iterations \citep{mattackv2,mpcattack,prafattack} and continue to improve even after 1,000 iterations \citep{prafattack}. Since each iteration requires forward and backward passes through several vision encoders, generating adversarial perturbations remains computationally expensive.

One way to reduce the computational cost of gradient-based attacks is to start from a perturbation that already contains useful information about the source and target images. Previous work has explored informed initialization for adversarial attacks. For image classifiers, \citet{nn-fgsm} use a trained generator to accelerate gradient-based attacks, although their approach requires gradients from the victim model. For VLMs, \citet{RevisitingCroPA2025} and \citet{attackvlm} use adversarial images generated by a separate gradient-based attack as their starting point. While these approaches demonstrate the potential of informed initialization, existing methods for VLMs still require costly gradient-based optimization to construct the initial perturbation. This motivates our research question: \textit{Can we learn to generate inexpensive adversarial perturbations that serve as effective warm starts for state-of-the-art gradient-based attacks?}


In this work, we study amortized initialization for transferable adversarial attacks on VLMs. Amortized adversarial generation allows a model to learn from many image pairs and produce adversarial perturbations for new inputs, without solving a separate gradient-based optimization problem each time. Unlike conventional gradient-based attacks, which start each optimization independently, a trained generator can use information learned from previous examples to produce an informed starting point. To study this approach, we introduce SPARK, a lightweight looped perturbation generator conditioned on both the source and target images. SPARK repeatedly applies the same network to refine a perturbation, with each step predicting an update to the perturbation produced by the previous step. We train SPARK to generate adversarial perturbations that are effective on their own, without explicitly optimizing them as initializations for another attack. At inference time, SPARK produces these perturbations in a few forward passes, without requiring gradients through the surrogate vision encoders. We then use the generated perturbations to initialize existing gradient-based attacks, which can further refine them using their original optimization procedures. Figure~\ref{fig:spark-overview} illustrates the pipeline, from amortized perturbation generation with SPARK to subsequent gradient-based refinement.

\begin{figure}[!b]
    \centering
    \includegraphics[width=1\linewidth]{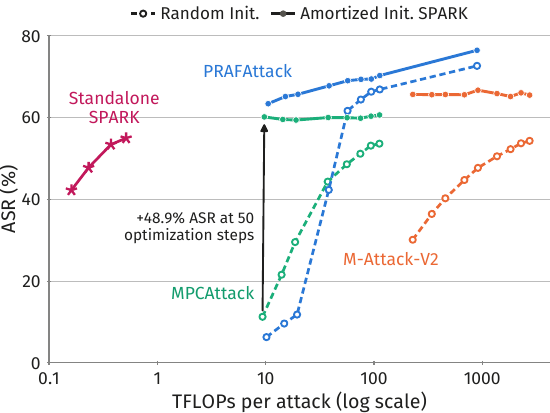}
    \caption{ASR versus TFLOPs per attack (log scale) for attacks initialized randomly or with SPARK. Points along each attack curve use increasing numbers of optimization steps. Standalone SPARK points use increasing numbers of generator calls.}
    \label{fig:compute-tradeoff}
\end{figure}

Our experiments show that SPARK initialization improves the attack success rates (ASR) of three state-of-the-art gradient-based methods: PRAFAttack \citep{prafattack}, MPCAttack \citep{mpcattack}, and MAttack-v2 \citep{mattackv2}. Averaged over six black-box VLMs, SPARK increases their ASR by 7.4, 11.4, and 15.3 percentage points, respectively, on ImageNet classes seen during generator training. Comparable improvements persist on held-out ImageNet classes and ObjectNet, showing that the benefits extend to unseen image categories. We also evaluate three frontier VLMs and find that SPARK improves the strongest five-encoder configuration of PRAFAttack by 4.4 percentage points, achieving new state-of-the-art ASR. Beyond improving attack success, SPARK also reduces the number of optimization steps required. With just 50 gradient-based steps, SPARK-initialized PRAFAttack achieves 63.4\% attack success, surpassing the 61.6\% obtained with 300 steps from random initialization. Even without subsequent gradient-based optimization, SPARK achieves 54.9\% attack success on ImageNet, outperforming MPCAttack while requiring approximately 100$\times$ less computation. Figure~\ref{fig:compute-tradeoff} compares ASR and computational cost for random and amortized initialization.

Our main contributions are summarized as follows:

\begin{itemize}
    \item We introduce SPARK, a lightweight looped perturbation generator that learns to produce adversarial perturbations from source-target image pairs in a few forward passes. These perturbations can be used to initialize existing gradient-based attacks without further modifications.

    \item We show that amortized initialization improves the attack success rates of three state-of-the-art gradient-based methods: PRAFAttack, MPCAttack, and MAttack-v2. These improvements generalize to unseen image categories and extend to frontier VLMs.

    \item We demonstrate that SPARK reduces the number of gradient-based optimization steps required to achieve high attack success rates. Even without subsequent optimization, SPARK produces effective standalone attacks at approximately 100$\times$ lower computational cost.
\end{itemize}

\section{Preliminaries and Related Work}
\label{sec:preliminaries}

We consider targeted transferable adversarial attacks against black-box vision-language models (VLMs). Given a source image $x_s$ and a target image $x_t$, the goal is to construct an adversarial image $x_s+\delta$ that causes a VLM to interpret the source image as depicting the target, while keeping the perturbation $\delta$ within an $\ell_\infty$ budget $\epsilon$. Since the victim VLM is inaccessible, transferable attacks instead optimize the perturbation using a set of surrogate vision encoders $\{E_i\}_{i=1}^M$. The underlying idea is to make the adversarial image resemble the target in the representation spaces of these encoders, even though the two images remain visually different. This can be formulated as
\begin{equation}
    \delta^* =
    \arg\max_{\|\delta\|_\infty \leq \epsilon}
    \frac{1}{M}\sum_{i=1}^{M}
    \operatorname{sim}\left(
        E_i(x_s+\delta), E_i(x_t)
    \right),
    \label{eq:transfer-objective}
\end{equation}
where $\operatorname{sim}(\cdot,\cdot)$ measures similarity between image representations. By optimizing across multiple surrogate encoders, the attack seeks perturbations that transfer to VLMs whose parameters and gradients are unavailable.

\noindent\textbf{Gradient-based attacks.}
Gradient-based transferable attacks optimize adversarial perturbations using gradients from surrogate vision encoders, with the aim of transferring them to black-box VLMs. Early work demonstrated that perturbations optimized on open-source models could transfer to proprietary VLMs \citep{dong2023robust}. Subsequent methods improved transferability by aligning representations across different image regions, transformations, and target references. MAttack \citep{mattack} introduced multi-crop feature alignment, while \citet{hu2025transferable} and M-Attack-V2 \citep{mattackv2} explored multiple target images and additional transformations. MPCAttack \citep{mpcattack} further improved transferability by jointly optimizing representations from diverse surrogate encoders. More recent methods exploit features beyond the final encoder layer. FOA-Attack \citep{jia2025foa} aligns spatial patches using optimal transport, while PRAFAttack \citep{prafattack} extends this approach with intermediate-layer alignment, progressive target scaling, and alternative patch-selection strategies. 

\noindent\textbf{Generator-based attacks.} An alternative is to learn a generator that produces adversarial images or perturbations from input images. Diffusion-based approaches such as AdvDiffVLM \citep{guo2024advdiffvlm} and PSI \citep{yao2026psi} use pretrained diffusion models, but still guide generation using gradients from surrogate encoders. In contrast, AnyAttack \citep{zhang2025anyattack} trains a neural network to directly generate adversarial perturbations, avoiding iterative gradient-based optimization at inference time. By learning across many image pairs, such generators can reuse information from training to produce perturbations for unseen inputs in a few forward passes. However, these approaches have primarily been studied as standalone attacks, rather than as initializations for gradient-based methods.

\textbf{Informed initialization for adversarial attacks.}
Informed initialization is a well-established approach for accelerating iterative optimization processes \citep{arthur2007kmeans, hinton2006fast}. \citet{finn2017model} demonstrated that initializations can themselves be learned to enable rapid adaptation through gradient descent. Prior work has also demonstrated its benefits for adversarial attacks. \citet{nn-fgsm} trained a generator to initialize perturbations for fast adversarial training. For VLMs, AttackVLM \citep{attackvlm} initializes black-box gradient-free optimization with adversarial images optimized on surrogate models. \citet{RevisitingCroPA2025} accelerate prompt-image co-optimization \citep{luo2024imageworth1000lies} by first optimizing the adversarial image independently using a gradient-based attack. Whether inexpensive learned initializations can further improve state-of-the-art transferable attacks, and whether a single learned initialization can generalize across gradient-based attacks with different optimization objectives, remains unexplored.

\section{SPARK: Amortized Adversarial Perturbation Generation}
\label{sec:method}

The optimization problem in Eq.~\eqref{eq:transfer-objective} is typically solved independently for each source-target image pair. Gradient-based attacks repeatedly update the perturbation using surrogate encoder gradients, without benefiting from perturbations optimized for previous image pairs. An alternative is to learn a function that maps a source-target pair directly to an adversarial perturbation. By training this function across many image pairs, we can amortize the cost of adversarial optimization. Instead of solving the optimization problem from scratch for every new pair, the model learns how to produce perturbations that align source and target representations.
In this section, we introduce SPARK, a lightweight looped generator that learns to produce adversarial perturbations from source-target image pairs. We explain how SPARK progressively refines perturbations, how it is trained, and how its outputs can be used as standalone attacks or to initialize existing gradient-based methods. Figure~\ref{fig:spark-overview} provides an overview of the approach.

\subsection{Learning to Generate Effective Adversarial Perturbations}
\label{sec:spark-generation}

SPARK learns to generate perturbations that align the source image with the target in the surrogate representation spaces, as described in Eq.~\eqref{eq:transfer-objective}. Given a source image $x_s$ and a target image $x_t$, we first extract the target's global representations using $M$ frozen surrogate vision encoders and concatenate them into a conditioning vector $z_t = [E_1(x_t), \ldots, E_M(x_t)]$. The generator receives this vector together with the source image and a current perturbation.
Rather than predicting the final perturbation in a single pass, SPARK treats perturbation generation as a learned iterative process. By learning how to update perturbations across many image pairs, SPARK amortizes the cost of adversarial optimization instead of solving each instance independently. A looped network repeatedly applies this learned update rule, with each step building on the previous perturbation. Sharing weights across steps encourages the network to learn a reusable refinement rule while keeping the number of trainable parameters fixed~\citep{dehghani2018universal}.

We implement this refinement using an unconstrained state $h^{(k)}$. Starting from a random perturbation $\delta^{(0)} \sim \mathcal{U}(-\epsilon,\epsilon)$, we initialize $h^{(0)} = \operatorname{arctanh}(\delta^{(0)}/\epsilon)$. At each refinement step, the generator takes the source image, target representations, normalized current perturbation, and step index to predict an update to the state
\begin{equation}
    h^{(k+1)} = h^{(k)} +
    G_\theta\!\left(x_s, z_t, \tanh(h^{(k)}), k\right),
    \label{eq:spark-state}
\end{equation}
where $G_\theta$ denotes the shared generator with parameters $\theta$. The updated state is then mapped to a bounded perturbation
\begin{equation}
    \delta^{(k+1)} = \epsilon \tanh\!\left(h^{(k+1)}\right).
    \label{eq:spark-perturbation}
\end{equation}
Since $\tanh$ maps each element to $(-1,1)$, scaling by the perturbation budget $\epsilon$ ensures that $\|\delta^{(k+1)}\|_\infty \leq \epsilon$ without requiring explicit projection.
We implement $G_\theta$ as an attention U-Net with a Transformer bottleneck and use $K=4$ refinement steps \citep{ozan2018attentionunet, chen2021transunet}. The U-Net processes the source image and current perturbation at multiple spatial resolutions, while the Transformer allows information to be exchanged across spatial locations. The target representations and refinement-step index condition the network through its convolutional blocks. Because the same network is used at every step, SPARK can perform multiple refinements without increasing the number of trainable parameters.

To train SPARK, we need a loss that measures whether its generated perturbations achieve the feature alignment sought in Eq.~\eqref{eq:transfer-objective}. Aligning only the final global image representations provides a useful signal, but does not capture all the information available in the surrogate encoders. Following PRAFAttack \citep{prafattack}, we therefore combine global feature alignment with alignment of selected intermediate-layer representations. 
For an adversarial image $x_a$ and target image $x_t$, the loss is
\begin{equation}
\begin{aligned}
    \mathcal{L}_{\mathrm{PRAF}}(x_a,x_t)
    &= -\mathcal{S}_{\mathrm{global}}(x_a,x_t)\\
    &\quad + 0.5\,\mathcal{L}_{\mathrm{CLS}}(x_a,x_t)\\
    &\quad + 1.5\,\mathcal{L}_{\mathrm{patch}}(x_a,x_t),
\end{aligned}
\label{eq:spark-praf-loss}
\end{equation}
where $\mathcal{S}_{\mathrm{global}}$ is the average cosine similarity between the final global representations of $x_a$ and $x_t$ across surrogate encoders. The remaining terms measure cosine distance between their corresponding \texttt{CLS} tokens and spatial patch tokens at selected intermediate layers. For patch alignment, we average over the 60\% of corresponding patches with the highest cosine similarity. Following PRAFAttack, intermediate layers are selected for each image pair based on the alignment of their input gradients with the final-layer gradient. This provides additional feature-alignment signals beyond the global representations while avoiding unreliable patch correspondences.

We supervise every refinement step to encourage SPARK to produce effective adversarial perturbations throughout the generation process, rather than only at the final step. The training objective averages the feature-alignment loss across all $K$ steps
\begin{equation}
    \mathcal{L}_{\mathrm{SPARK}}(\theta)
    =
    \frac{1}{K}\sum_{k=1}^{K}
    \mathcal{L}_{\mathrm{PRAF}}
    \left(
        A_k(x_s+\delta^{(k)}),\,x_t
    \right),
    \label{eq:spark-training}
\end{equation}
where $\delta^{(k)}$ is the perturbation generated at step $k$ and $A_k$ denotes a randomly sampled image crop. During training, gradients propagate through the frozen surrogate encoders and the looped generator to update its parameters $\theta$. At inference time, SPARK generates perturbations using only forward passes, without requiring gradients through the surrogate encoders.

\subsection{Using Generated Perturbations to Improve Gradient-Based Attacks}
\label{sec:spark-initialization}

SPARK is trained to generate effective adversarial perturbations without relying on subsequent gradient-based optimization. However, the perturbations it produces can also serve as informed starting points for existing attacks. Unlike random initialization, which contains no information about the source-target pair, SPARK's output has already been optimized through learning to align their surrogate representations. This suggests that gradient-based attacks can build on the learned perturbation rather than starting the optimization from scratch.

To use SPARK as an initialization, we simply replace the random starting perturbation of a gradient-based attack with the output of the trained generator,
\begin{equation}
    \delta_{\mathrm{init}} = \delta^{(K)} = \mathrm{SPARK}_{\theta}(x_s,x_t).
    \label{eq:spark-initialization}
\end{equation}
The gradient-based attack then refines $\delta_{\mathrm{init}}$ using its original objective and optimization procedure, without further modifications. This allows SPARK to initialize different attacks, even when their optimization objectives differ from the PRAFAttack loss used to train the generator. Since generating the initial perturbation requires only a few forward passes, the additional computational cost is small compared with hundreds of gradient-based optimization steps.

\section{Experimental Setup}
\label{sec:experimental_setup}

We evaluate SPARK as both a standalone adversarial attack and an initialization for existing gradient-based attacks. Our experiments study attack effectiveness, computational efficiency, and generalization across image categories, surrogate encoders, and black-box VLMs. We first describe the datasets, attack baselines, and evaluation protocol used throughout our experiments.

\textbf{Datasets.} We construct training and evaluation datasets from ImageNet \citep{russakovsky2015imagenet}. To evaluate generalization to unseen image categories, we randomly split the 1,000 ImageNet classes into 800 in-distribution (ID) classes used for SPARK training and 200 out-of-distribution (OOD) classes held out from training. Within each group, we establish a fixed one-to-one mapping between source and target classes, excluding self-pairs. Each image pair consists of a source image to perturb and a target image that specifies the desired adversarial content. We construct 50,400 training pairs from ImageNet-train, with 63 pairs per ID class mapping.

For evaluation, we use disjoint images from ImageNet-val to construct ID validation, ID test, and OOD test sets, each containing 1,000 image pairs. The ID sets measure performance on new images from classes seen during training, while the OOD set tests whether SPARK generalizes to entirely unseen classes. We additionally evaluate on 1,000 ObjectNet \citep{barbu2019objectnet} examples, excluding classes used for ID training. ObjectNet provides a complementary test of transferability to images from a different dataset.

\noindent\textbf{Attack baselines.}
We evaluate whether SPARK initialization improves three state-of-the-art gradient-based transferable attacks: PRAFAttack \citep{prafattack}, MPCAttack \citep{mpcattack}, and MAttack-v2 \citep{mattackv2}. To ensure a consistent comparison, all three attacks use the same surrogate vision encoders employed to train SPARK: CLIP ViT-B/16 \citep{clip}, LAION OpenCLIP ViT-B/32 \citep{openclip}, and DINOv2 ViT-B/14 \citep{oquab2024dinov2learningrobustvisual}. For each method, we compare random initialization against initialization with a perturbation generated by SPARK. We use 300 gradient-based optimization steps for the main comparisons and extend the budget to 600 steps to study convergence. We also compare SPARK as a standalone generative attack against AnyAttack \citep{zhang2025anyattack} and existing gradient-based methods, considering both attack success and computational cost. Finally, to test whether SPARK generalizes across surrogate ensembles, we evaluate its ability to initialize the stronger five-encoder configuration of PRAFAttack without retraining the generator.

\textbf{Black-box evaluation.}
We evaluate attacks against six open-source VLMs in a black-box setting: Gemma 4 12B \citep{team2026gemma}, Qwen3-VL 8B \citep{bai2025qwen3}, InternVL3.5 8B \citep{wang2025internvl3}, GLM-4.6V Flash \citep{glm}, Kimi-VL A3B \citep{team2025kimi}, and Phi-4-reasoning-vision 15B \citep{aneja2026phi}. We additionally evaluate three frontier VLMs, GPT-6.1 Sol \citep{openai2026gpt61sol}, Claude Opus 5.5 \citep{anthropic2026opus55}, and Gemini 4 Argon \citep{google2026gemini4argon}, in selected experiments. We disable reasoning where supported and otherwise use low reasoning effort.

We follow the GPTScore-based evaluation protocol of prior transferable attacks \citep{mattackv2,mpcattack,prafattack}. For each image pair, we prompt the victim VLM to generate short captions of the target image $x_t$ and adversarial image $x_a$. We then use Qwen-3.8-27B \citep{qwen2026qwen38} as an LLM judge to assess their semantic similarity, assigning a score between 0 and 1. We report the average GPTScore and ASR, defined as the percentage of image pairs with a similarity score of at least $0.5$.

\providecommand{\dcell}[1]{{\footnotesize\textit{#1}}}
\begin{table*}[t]
    \centering
    \renewcommand{\arraystretch}{1.2}
    \setlength{\aboverulesep}{0pt}\setlength{\belowrulesep}{0pt}\setlength{\extrarowheight}{1pt}
    \caption{Per-victim ASR (\%) and average GPTScore for random vs. amortized initialization of gradient-based attacks. MAV2 denotes M-Attack-V2 \citep{mattackv2}, MPC denotes MPCAttack \citep{mpcattack}, and PRAF denotes PRAFAttack \citep{prafattack}. All attacks are run for 300 optimization steps. $\mathrm{G_\theta}+$ denotes amortized initialization using a $K=4$ Generator $\mathrm{G_\theta}$.}
    \label{tab:main-table}
    \resizebox{\textwidth}{!}{%
    \begin{tabular}{l cc cc cc cc cc cc | cc}
    \toprule
     & \multicolumn{2}{c}{Gemma-4-12B} & \multicolumn{2}{c}{Qwen3-VL-8B} & \multicolumn{2}{c}{InternVL3.5-8B} & \multicolumn{2}{c}{GLM-4.6V-Flash} & \multicolumn{2}{c}{Kimi-VL-A3B} & \multicolumn{2}{c|}{Phi-4-RV-15B} & \multicolumn{2}{c}{Avg.} \\
    \cmidrule(lr){2-3} \cmidrule(lr){4-5} \cmidrule(lr){6-7} \cmidrule(lr){8-9} \cmidrule(lr){10-11} \cmidrule(lr){12-13} \cmidrule(lr){14-15}
    Method & ASR & AvgScore & ASR & AvgScore & ASR & AvgScore & ASR & AvgScore & ASR & AvgScore & ASR & AvgScore & ASR & AvgScore \\
    \midrule
    \multicolumn{15}{c}{\texttt{ID-test}} \\
    \midrule
    \rowcolor{white!5} MAttackV2 & 54.0 & 0.508 & 56.9 & 0.515 & 55.2 & 0.497 & 53.9 & 0.484 & 49.9 & 0.458 & 33.1 & 0.357 & 50.5 & 0.470 \\
    \rowcolor{white!5} \textcolor{magenta}{+ SPARK} & \underline{70.4} & \underline{0.601} & \underline{73.7} & \underline{0.622} & \underline{67.2} & \underline{0.582} & \underline{67.9} & \underline{0.569} & \underline{66.7} & \underline{0.553} & \underline{49.0} & \underline{0.450} & \underline{65.8} & \underline{0.563} \\
    \rowcolor{blue!5} \quad\textit{$\Delta$} & \dcell{+16.4} & \dcell{+0.093} & \dcell{+16.8} & \dcell{+0.107} & \dcell{+12.0} & \dcell{+0.085} & \dcell{+14.0} & \dcell{+0.085} & \dcell{+16.8} & \dcell{+0.095} & \dcell{+15.9} & \dcell{+0.093} & \dcell{+15.3} & \dcell{+0.093} \\
    \rowcolor{white!5} MPCAttack & 53.5 & 0.501 & 55.5 & 0.512 & 51.5 & 0.482 & 48.5 & 0.451 & 48.0 & 0.453 & 34.3 & 0.361 & 48.5 & 0.460 \\
    \rowcolor{white!5} \textcolor{magenta}{+ SPARK} & 64.5 & 0.566 & 70.3 & 0.591 & 62.7 & 0.546 & 58.5 & 0.511 & 60.9 & 0.526 & 42.7 & 0.413 & 59.9 & 0.526 \\
    \rowcolor{blue!5} \quad\textit{$\Delta$} & \dcell{+11.0} & \dcell{+0.065} & \dcell{+14.8} & \dcell{+0.079} & \dcell{+11.2} & \dcell{+0.064} & \dcell{+10.0} & \dcell{+0.060} & \dcell{+12.9} & \dcell{+0.073} & \dcell{+8.4} & \dcell{+0.052} & \dcell{+11.4} & \dcell{+0.066} \\
    \rowcolor{white!5} PRAFAttack & 67.1 & 0.582 & 67.9 & 0.585 & 65.4 & 0.569 & 64.1 & 0.549 & 61.3 & 0.524 & 43.8 & 0.414 & 61.6 & 0.537 \\
    \rowcolor{white!5} \textcolor{magenta}{+ SPARK} & \textbf{71.9} & \textbf{0.615} & \textbf{76.6} & \textbf{0.647} & \textbf{72.5} & \textbf{0.608} & \textbf{71.3} & \textbf{0.604} & \textbf{68.2} & \textbf{0.575} & \textbf{53.3} & \textbf{0.480} & \textbf{69.0} & \textbf{0.588} \\
    \rowcolor{blue!5} \quad\textit{$\Delta$} & \dcell{+4.8} & \dcell{+0.033} & \dcell{+8.7} & \dcell{+0.062} & \dcell{+7.1} & \dcell{+0.039} & \dcell{+7.2} & \dcell{+0.055} & \dcell{+6.9} & \dcell{+0.051} & \dcell{+9.5} & \dcell{+0.066} & \dcell{+7.4} & \dcell{+0.051} \\
    \midrule
    \multicolumn{15}{c}{\texttt{OOD-test}} \\
    \midrule
    \rowcolor{white!5} MAttackV2 & 53.2 & 0.495 & 57.2 & 0.514 & 53.4 & 0.489 & 51.3 & 0.465 & 49.2 & 0.448 & 33.8 & 0.357 & 49.7 & 0.461 \\
    \rowcolor{white!5} \textcolor{magenta}{+ SPARK} & \underline{68.4} & \underline{0.589} & \underline{74.7} & \underline{0.621} & \underline{67.7} & \underline{0.581} & \underline{65.7} & \underline{0.564} & \textbf{63.2} & \underline{0.540} & \underline{47.2} & \underline{0.447} & \underline{64.5} & \underline{0.557} \\
    \rowcolor{blue!5} \quad\textit{$\Delta$} & \dcell{+15.2} & \dcell{+0.094} & \dcell{+17.5} & \dcell{+0.107} & \dcell{+14.3} & \dcell{+0.092} & \dcell{+14.4} & \dcell{+0.099} & \dcell{+14.0} & \dcell{+0.092} & \dcell{+13.4} & \dcell{+0.090} & \dcell{+14.8} & \dcell{+0.096} \\
    \rowcolor{white!5} MPCAttack & 49.1 & 0.470 & 51.5 & 0.484 & 48.1 & 0.463 & 42.9 & 0.421 & 45.9 & 0.429 & 30.5 & 0.343 & 44.7 & 0.435 \\
    \rowcolor{white!5} \textcolor{magenta}{+ SPARK} & 63.6 & 0.548 & 67.3 & 0.576 & 60.1 & 0.526 & 56.6 & 0.505 & 58.6 & 0.508 & 40.7 & 0.402 & 57.8 & 0.511 \\
    \rowcolor{blue!5} \quad\textit{$\Delta$} & \dcell{+14.5} & \dcell{+0.078} & \dcell{+15.8} & \dcell{+0.092} & \dcell{+12.0} & \dcell{+0.063} & \dcell{+13.7} & \dcell{+0.084} & \dcell{+12.7} & \dcell{+0.079} & \dcell{+10.2} & \dcell{+0.059} & \dcell{+13.1} & \dcell{+0.076} \\
    \rowcolor{white!5} PRAFAttack & 63.9 & 0.558 & 66.8 & 0.577 & 63.5 & 0.557 & 60.9 & 0.535 & 55.7 & 0.499 & 41.9 & 0.413 & 58.8 & 0.523 \\
    \rowcolor{white!5} \textcolor{magenta}{+ SPARK} & \textbf{71.5} & \textbf{0.607} & \textbf{76.3} & \textbf{0.640} & \textbf{68.9} & \textbf{0.593} & \textbf{69.8} & \textbf{0.586} & \underline{63.1} & \textbf{0.549} & \textbf{49.8} & \textbf{0.460} & \textbf{66.6} & \textbf{0.572} \\
    \rowcolor{blue!5} \quad\textit{$\Delta$} & \dcell{+7.6} & \dcell{+0.049} & \dcell{+9.5} & \dcell{+0.063} & \dcell{+5.4} & \dcell{+0.036} & \dcell{+8.9} & \dcell{+0.051} & \dcell{+7.4} & \dcell{+0.050} & \dcell{+7.9} & \dcell{+0.047} & \dcell{+7.8} & \dcell{+0.049} \\
    \midrule
    \multicolumn{15}{c}{\texttt{ObjectNet-test}} \\
    \midrule
    \rowcolor{white!5} MAttackV2 & 33.8 & 0.396 & 38.8 & 0.425 & 32.6 & 0.381 & 31.1 & 0.366 & 28.3 & 0.349 & 19.8 & 0.290 & 30.7 & 0.368 \\
    \rowcolor{white!5} \textcolor{magenta}{+ SPARK} & \underline{52.4} & \underline{0.502} & \underline{57.8} & \underline{0.527} & \underline{44.9} & \underline{0.454} & \underline{43.6} & 0.443 & \underline{45.1} & \underline{0.442} & 28.8 & 0.345 & \underline{45.4} & \underline{0.452} \\
    \rowcolor{blue!5} \quad\textit{$\Delta$} & \dcell{+18.6} & \dcell{+0.106} & \dcell{+19.0} & \dcell{+0.102} & \dcell{+12.3} & \dcell{+0.073} & \dcell{+12.5} & \dcell{+0.077} & \dcell{+16.8} & \dcell{+0.093} & \dcell{+9.0} & \dcell{+0.055} & \dcell{+14.7} & \dcell{+0.084} \\
    \rowcolor{white!5} MPCAttack & 33.4 & 0.401 & 40.9 & 0.430 & 31.3 & 0.371 & 31.0 & 0.364 & 30.9 & 0.367 & 22.0 & 0.306 & 31.6 & 0.373 \\
    \rowcolor{white!5} \textcolor{magenta}{+ SPARK} & 45.5 & 0.464 & 51.0 & 0.492 & 38.3 & 0.415 & 40.5 & 0.414 & 40.8 & 0.420 & 29.5 & 0.346 & 40.9 & 0.425 \\
    \rowcolor{blue!5} \quad\textit{$\Delta$} & \dcell{+12.1} & \dcell{+0.063} & \dcell{+10.1} & \dcell{+0.062} & \dcell{+7.0} & \dcell{+0.044} & \dcell{+9.5} & \dcell{+0.050} & \dcell{+9.9} & \dcell{+0.053} & \dcell{+7.5} & \dcell{+0.040} & \dcell{+9.3} & \dcell{+0.052} \\
    \rowcolor{white!5} PRAFAttack & 48.6 & 0.482 & 53.1 & 0.508 & 41.3 & 0.438 & 43.5 & \underline{0.444} & 41.3 & 0.426 & \underline{29.9} & \underline{0.356} & 43.0 & 0.442 \\
    \rowcolor{white!5} \textcolor{magenta}{+ SPARK} & \textbf{56.3} & \textbf{0.533} & \textbf{63.3} & \textbf{0.564} & \textbf{51.9} & \textbf{0.493} & \textbf{55.0} & \textbf{0.503} & \textbf{51.4} & \textbf{0.481} & \textbf{34.1} & \textbf{0.379} & \textbf{52.0} & \textbf{0.492} \\
    \rowcolor{blue!5} \quad\textit{$\Delta$} & \dcell{+7.7} & \dcell{+0.051} & \dcell{+10.2} & \dcell{+0.056} & \dcell{+10.6} & \dcell{+0.055} & \dcell{+11.5} & \dcell{+0.059} & \dcell{+10.1} & \dcell{+0.055} & \dcell{+4.2} & \dcell{+0.023} & \dcell{+9.0} & \dcell{+0.050} \\
    \bottomrule
    \end{tabular}}
\end{table*}

\section{Results}
\label{sec:results}

In this section, we evaluate the effect of amortized initialization using SPARK on existing gradient-based attacks, SPARK as a standalone attack, and generalization to stronger ensembles and frontier black-box VLMs.

\subsection{Amortized Initialization Improves Gradient-Based Attacks}

Table~\ref{tab:main-table} compares random and SPARK initialization for three gradient-based attacks across six black-box VLMs. On \texttt{ID-test}, SPARK improves the average attack success rates of PRAFAttack, MPCAttack, and MAttack-v2 by 7.4, 11.4, and 15.3 percentage points, respectively. Notably, these improvements occur despite differences in the attacks' optimization objectives. PRAFAttack remains the strongest method overall, reaching 69.0\% average ASR with SPARK initialization. Since SPARK is trained using the PRAFAttack loss, its ability to improve MPCAttack and MAttack-v2 also shows that the learned perturbations are useful beyond the objective used during generator training.

These improvements extend to image categories excluded from SPARK training. On \texttt{OOD-test}, SPARK provides gains comparable to those observed on \texttt{ID-test}, despite the source and target classes being entirely unseen during training. The gains also persist on ObjectNet, where the images come from a different dataset. In fact, SPARK-initialized PRAFAttack achieves a larger improvement on \texttt{ObjectNet-test} than on \texttt{ID-test}, increasing average ASR by 9.0 percentage points. This suggests that the benefit of amortized initialization is not limited to the image categories or dataset used to train the generator.

\begin{figure*}[t]
    \centering
    \begin{subfigure}[c]{0.75\textwidth}
        \centering
        \includegraphics[width=\linewidth]{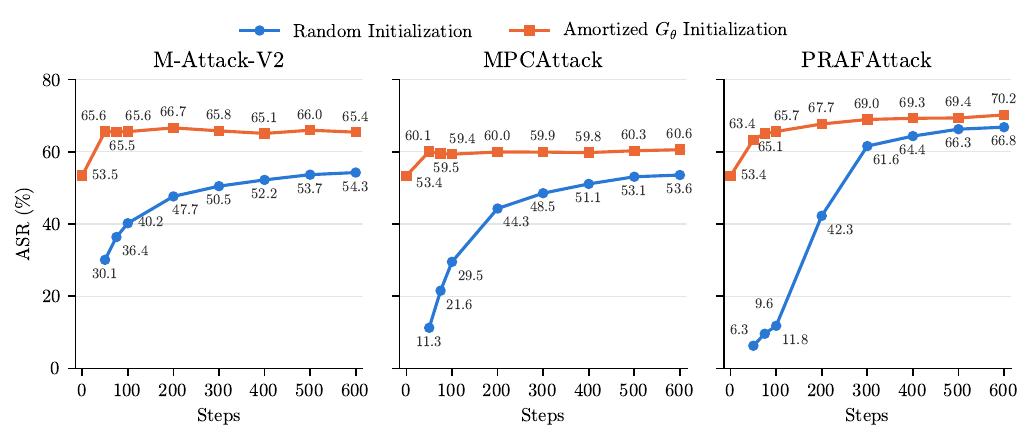}
    \end{subfigure}
    \hfill
    \begin{subfigure}[c]{0.24\textwidth}
        \centering
        \includegraphics[width=\linewidth]{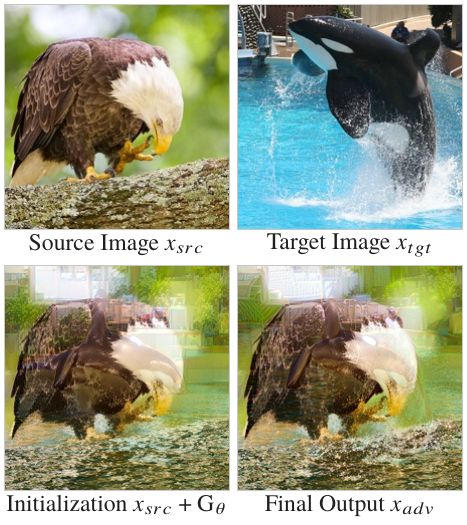}
    \end{subfigure}
    \vspace{-1em}
    \caption{Left) Average ASR (\%) vs. attack steps for random and amortized initialization on \texttt{ID-test}. Right) Qualitative example of amortized initialization of PRAFAttack. We show additional qualitative examples in Appendix \ref{app:qualitative-examples}.}
    \label{fig:asr-vs-steps}
\end{figure*}

\subsection{Attack Success and Computational Cost}

Figure~\ref{fig:asr-vs-steps} shows how attack success changes as we increase the number of gradient-based optimization steps. SPARK initialization provides a substantial advantage early in optimization, but the gains also persist at larger budgets. Even after 600 steps, SPARK-initialized PRAFAttack achieves 70.2\% ASR on \texttt{ID-test}, compared with 66.8\% from random initialization. While MPCAttack and MAttack-v2 approach saturation relatively quickly with SPARK initialization, PRAFAttack continues to improve with additional optimization. This shows that SPARK provides more than an initial acceleration, as its benefits persist even after extensive gradient-based refinement.

SPARK also substantially reduces the optimization required to achieve high attack success. With just 50 gradient-based steps, SPARK-initialized PRAFAttack achieves 63.4\% ASR, surpassing the 61.6\% obtained with 300 steps from random initialization. This reduces the number of gradient-based steps by a factor of six, while adding only a few forward passes to generate the initial perturbation. Before any gradient-based refinement, SPARK already achieves 53.4\% ASR with four looped passes, which increases by 10.0 percentage points after 50 PRAFAttack steps. These results show how amortized generation and gradient-based optimization complement each other, combining inexpensive perturbation generation with subsequent refinement.

\subsection{Standalone Adversarial Generation}

SPARK can also generate adversarial perturbations without subsequent gradient-based optimization. Table~\ref{tab:generalization} compares its standalone performance with existing gradient-based and generative attacks. With $K=4$ looped steps, SPARK achieves 53.4\% ASR on \texttt{ID-test}, outperforming MPCAttack and MAttack-v2 after 300 optimization steps while requiring approximately $150\times$ less computation than MPCAttack. SPARK also substantially outperforms AnyAttack \citep{zhang2025anyattack}, although the two methods differ in their conditioning, as AnyAttack does not use the source image as an input. These results show that amortized generation can produce effective adversarial perturbations without iterative gradient optimization at inference time.

Increasing the number of looped steps consistently improves standalone attack success across all three datasets. On \texttt{ID-test}, ASR increases from 42.2\% with $K=1$ to 54.9\% with $K=6$, suggesting that repeated refinement helps the generator construct stronger perturbations. Each additional step requires only another forward pass through the shared network, without introducing new parameters. The total inference cost is approximately $0.090 + 0.071K$ TFLOPs per image, including target encoding and perturbation generation. Although standalone performance decreases on unseen ImageNet classes and ObjectNet, SPARK remains effective without any inference-time gradient-based optimization.


\begin{table}[ht]
    \centering
    \caption{Generalization across test sets (1,000 image pairs each). MPCAttack and PRAFAttack run 300 optimization steps; $K$ is the number of SPARK's iterations during training and inference.}
    \label{tab:generalization}
    \resizebox{\linewidth}{!}{%
    \begin{tabular}{l@{\hspace{3pt}}r cc cc cc}
    \toprule
     & 
    \multirow{2}{*}{%
        \makebox[0pt][r]{%
            \footnotesize $\displaystyle\frac{\mathrm{TFLOPs}}{\mathrm{image}}$%
        }%
    }
         
     & \multicolumn{2}{c}{ID Test} & \multicolumn{2}{c}{OOD Test} & \multicolumn{2}{c}{ObjectNet Test} \\
    \cmidrule(lr){3-4} \cmidrule(lr){5-6} \cmidrule(lr){7-8}
    Method & & \footnotesize{ASR} & \footnotesize{AvgScore} & \footnotesize{ASR} & \footnotesize{AvgScore} & \footnotesize{ASR} & \footnotesize{AvgScore} \\
    \midrule
    \multicolumn{8}{c}{\textit{Gradient-based Attacks}} \\
    \midrule
    MAttackV2 & 1366$^\dagger$ & 50.5 & 0.47 & \underline{49.7} & \underline{0.46} & 30.7 & 0.37 \\
    MPCAttack & 56.1 & 48.5 & 0.46 & 44.7 & 0.44 & \underline{31.6} & \underline{0.37} \\
    PRAFAttack & 57.0 & \textbf{61.6} & \textbf{0.54} & \textbf{58.8} & \textbf{0.52} & \textbf{43.0} & \textbf{0.44} \\
    \midrule
    \multicolumn{8}{c}{\textit{Generative Attacks}} \\
    \midrule
    AnyAttack & 0.013 & 0.4 & 0.05 & 0.3 & 0.05 & 1.2 & 0.11 \\
    $\mathrm{G_\theta}$ $K{=}1$ & 0.161 & 42.2 & 0.42 & 31.6 & 0.35 & 14.2 & 0.25 \\
    $\mathrm{G_\theta}$ $K{=}2$ & 0.232 & 47.7 & 0.45 & 35.3 & 0.37 & 15.4 & 0.26 \\
    $\mathrm{G_\theta}$ $K{=}4$ & 0.373 & 53.4 & 0.48 & 40.8 & 0.41 & 19.9 & 0.29 \\
    $\mathrm{G_\theta}$ $K{=}6$ & 0.515 & \underline{54.9} & \underline{0.50} & 42.8 & 0.43 & 20.7 & 0.30 \\
    \bottomrule
    \end{tabular}}
    \par\vspace{2pt}{\footnotesize\raggedright $^\dagger$M-Attack-V2 uses 10 crops of the source image and augments the primary target and 2 retrieved target images per iteration.\par}
\end{table}

\subsection{Generalization Across Surrogate Ensembles}

Our previous experiments use the same three surrogate encoders for SPARK and the gradient-based attacks to ensure consistent comparisons. However, stronger attacks can benefit from larger surrogate ensembles, and retraining SPARK for every configuration would be costly. We therefore test whether SPARK, trained with three encoders, can initialize the strongest five-encoder configuration of PRAFAttack \citep{prafattack}. This configuration uses OpenAI CLIP ViT-B/16 and ViT-B/32, LAION CLIP ViT-g/14, DINOv2-base, and the InternVL3-1B vision encoder. Importantly, we use the same trained SPARK generator without modification.

Table~\ref{tab:praf5-asr} shows that SPARK initialization improves five-encoder PRAFAttack across all three test datasets and all nine victim VLMs. The improvements extend to the three frontier VLMs, where average ASR increases by 4.4, 6.0, and 4.1 percentage points on \texttt{ID-test}, \texttt{OOD-test}, and \texttt{ObjectNet-test}, respectively. The gains are also consistent across the six open-source VLMs, despite the stronger attack already achieving higher success rates than its three-encoder counterpart. These results suggest that SPARK learns perturbations useful beyond the surrogate ensemble and optimization objective used during training, without requiring a generator for each attack configuration.

\providecommand{\dcell}[1]{{\footnotesize\textit{#1}}}
\providecommand{\vlm}[1]{{\footnotesize #1}}
\begin{table}[ht]
    \centering
    \renewcommand{\arraystretch}{1.2}
    \setlength{\tabcolsep}{3pt}
    \caption{ Amortized initialization under surrogate ensemble mismatch. Each cell shows ASR (\%). P denotes randomly initialized PRAFAttack. S$+$P denotes PRAFAttack initialized with SPARK. }
    \label{tab:praf5-asr}
    \resizebox{\linewidth}{!}{%
    \begin{tabular}{l cccccc|c ccc|c}
    \toprule
    Method & \multicolumn{7}{c}{Open-Source VLMs} & \multicolumn{4}{c}{Closed-Source VLMs} \\
    \cmidrule(lr){2-8} \cmidrule(lr){9-12}
     & \vlm{Gemm.} & \vlm{Qwen} & \vlm{Intern} & \vlm{GLM} & \vlm{Kimi} & \vlm{Phi} & \vlm{Avg.} & \vlm{GPT} & \vlm{Claude} & \vlm{Gemini} & \vlm{Avg.} \\
    \midrule
    \multicolumn{12}{c}{\texttt{ID-Test}} \\
    \midrule
    \rowcolor{white!4} P & 77.9 & 79.4 & 79.2 & 74.0 & 70.0 & 55.1 & 72.6 & 31.9 & 69.1 & 45.5 & 48.8 \\
    \rowcolor{white!4} S$+$P  & 83.4 & 84.0 & 82.2 & 77.1 & 73.5 & 57.3 & 76.2 & 36.8 & 72.1 & 50.7 & 53.2 \\
    \rowcolor{blue!10} \quad\textit{$\Delta$} & \dcell{+5.5} & \dcell{+4.6} & \dcell{+3.0} & \dcell{+3.1} & \dcell{+3.5} & \dcell{+2.2} & \dcell{+3.7} & \dcell{+4.9} & \dcell{+3.0} & \dcell{+5.2} & \dcell{+4.4} \\
    \midrule
    \multicolumn{12}{c}{\texttt{OOD-Test}} \\
    \midrule
    \rowcolor{white!4} P & 77.8 & 76.6 & 77.0 & 69.6 & 65.4 & 50.8 & 69.5 & 30.4 & 64.7 & 41.6 & 45.6 \\
    \rowcolor{white!4} S$+$P & 80.5 & 83.2 & 81.5 & 75.7 & 71.5 & 53.4 & 74.3 & 37.6 & 70.0 & 47.0 & 51.5 \\
    \rowcolor{blue!10} \quad\texttt{$\Delta$} & \dcell{+2.7} & \dcell{+6.6} & \dcell{+4.5} & \dcell{+6.1} & \dcell{+6.1} & \dcell{+2.6} & \dcell{+4.8} & \dcell{+7.2} & \dcell{+5.3} & \dcell{+5.4} & \dcell{+6.0} \\
    \midrule
    \multicolumn{12}{c}{\texttt{ObjectNet-Test}} \\
    \midrule
    \rowcolor{white!4} P & 62.9 & 64.4 & 58.6 & 52.7 & 49.2 & 34.1 & 53.6 & 16.4 & 45.9 & 28.3 & 30.2 \\
    \rowcolor{white!4} S$+$P & 70.3 & 72.3 & 63.7 & 59.9 & 54.4 & 36.9 & 59.6 & 17.3 & 49.6 & 36.1 & 34.3 \\
    \rowcolor{blue!10} \quad\textit{$\Delta$} & \dcell{+7.4} & \dcell{+7.9} & \dcell{+5.1} & \dcell{+7.2} & \dcell{+5.2} & \dcell{+2.8} & \dcell{+5.9} & \dcell{+0.9} & \dcell{+3.7} & \dcell{+7.8} & \dcell{+4.1} \\
    \bottomrule
    \end{tabular}}
\end{table}

\section{Ablations}
\label{sec:ablations}

We conduct ablation studies to understand which design choices contribute to SPARK's effectiveness. In particular, we investigate how the training objective, looped architecture, and perturbation budget affect the transferability of generated attacks. These experiments help distinguish the benefits of learned iterative refinement from those of simply increasing model size or computational cost.

\textbf{Training Objective and Intermediate Supervision.}
We study the training objective and the importance of supervising intermediate refinement steps. When training only on the final output, ASR on \texttt{ID-test} decreases from 53.4\% to 52.2\% for $K=4$, with a larger reduction from 54.9\% to 52.5\% for $K=6$. This suggests that intermediate supervision becomes more useful as the number of refinement steps increases. Replacing the PRAFAttack-based training objective with the InfoNCE loss used by MPCAttack reduces \texttt{ID-test} ASR from 53.4\% to 41.6\% at $K=4$, with similar trends on the other datasets. These results suggest that the richer feature alignment provided by PRAFAttack is important for training effective transferable perturbations. Additional results are provided in Appendix~\ref{app:training-objective-ablation}.

\textbf{Architecture Ablations.}
We investigate whether SPARK benefits specifically from its looped, weight-shared architecture. Replacing the shared network with separate parameters for each of the four steps increases the model size from 9.8M to 39.1M parameters, yet reduces \texttt{ID-test} ASR from 53.4\% to 50.8\%. Interestingly, the larger model achieves lower training and validation losses but worse black-box transfer, suggesting that weight sharing may help prevent overfitting to the surrogate encoders. To separate the benefits of refinement from those of additional computation, we compare SPARK with a wider single-pass generator requiring comparable inference compute. Despite using similar FLOPs, the looped generator achieves 53.4\% ASR compared with 43.7\% for the wider model. Removing the Transformer bottleneck reduces single-pass ASR from 42.2\% to 31.2\%, despite only a modest reduction in computational cost. Together, these experiments suggest that iterative refinement, weight sharing, and spatial information exchange each contribute to SPARK's effectiveness. Full results are provided in Appendix~\ref{app:architecture-ablation}.

\textbf{Perturbation Budget Ablations.}
We test whether a generator trained with a perturbation budget of $\epsilon=16/255$ can be adapted to the smaller budget $\epsilon=8/255$ without retraining. Simply rescaling or clipping the generated perturbations performs poorly, achieving at most 5.3\% ASR on \texttt{ID-test}. In contrast, training SPARK directly with $\epsilon=8/255$ achieves 22.1\% ASR, comparable to PRAFAttack's 22.5\% under the same budget. SPARK initialization also improves PRAFAttack at this smaller budget, increasing ASR to 25.4\%. This suggests that the generator must learn perturbations for the intended budget, but the benefits of amortized initialization extend to smaller perturbations. Detailed results are provided in Appendix~\ref{app:perturbation-budget-ablation}.

\section{Conclusion}

In this work, we studied whether amortized adversarial generation can provide better starting points for gradient-based transferable attacks on black-box VLMs. We introduced SPARK, a lightweight looped generator that learns to produce adversarial perturbations from source-target image pairs. Our experiments show that these perturbations improve existing gradient-based attacks while substantially reducing the number of optimization steps needed to achieve high attack success. Importantly, SPARK improves attacks with different optimization objectives and remains effective when used with a stronger surrogate ensemble without retraining. These findings suggest that learned adversarial generators need not replace gradient-based optimization, but can instead complement it by providing effective initializations that transfer across attack methods.
\section{Acknowledgment}
YD acknowledges funding from the National Science Foundation under Grant No. 2431569. NBE would like to acknowledge support from DSO National Laboratories. 

{
    \small
    \bibliographystyle{ieeenat_fullname}
    \bibliography{main}
}

\clearpage

\onecolumn
\setcounter{page}{1}
{\centering
    \Large
    \textbf{\thetitle}\par
    \vspace{0.5em}
    Supplementary Material\par
}
\vspace{1.0em}

\section*{Appendix Contents}
{
    \small
    \setlength{\parskip}{6pt}
    \newcommand{\appendixcontentline}[2]{%
        \noindent\hspace*{#1}%
        \hyperref[#2]{\ref*{#2}\quad\nameref*{#2}}%
        \nobreak\dotfill\nobreak
        \hyperref[#2]{\pageref*{#2}}\par
    }
    \appendixcontentline{0em}{app:training-details}
    \appendixcontentline{1.5em}{app:training-compute}
    \appendixcontentline{1.5em}{app:loss-curves}
    \appendixcontentline{0em}{app:detailed-results}
    \appendixcontentline{1.5em}{app:asr-vs-steps-ood-objectnet}
    \appendixcontentline{1.5em}{app:intermediate_step_results_amortized_praf5}
    \appendixcontentline{1.5em}{app:gptscore-thresholds}
    \appendixcontentline{1.5em}{app:statistical-significance}
    \appendixcontentline{1.5em}{app:gptscore-judge}
    \appendixcontentline{0em}{app:ablations}
    \appendixcontentline{1.5em}{app:training-objective-ablation}
    \appendixcontentline{1.5em}{app:architecture-ablation}
    \appendixcontentline{1.5em}{app:perturbation-budget-ablation}
    \appendixcontentline{0em}{sec:inference-depth-unrolling}
    \appendixcontentline{0em}{app:qualitative-examples}
    \appendixcontentline{0em}{app:vlm-response-examples}
}

\clearpage

\section{SPARK Training Details}
\label{app:training-details}

\subsection{Computational Cost of Training}
\label{app:training-compute}

We estimate the cost of training SPARK analytically from its components. Target representations are precomputed once per target and cached, so each training sample requires $K$ generator calls and $K$ evaluations of the alignment loss (one per supervised refinement step). A generator forward pass costs $C_G \approx 0.071$ TFLOPs, and its backward pass costs approximately twice that, because the generator's parameters are trained. A forward pass through the three surrogate encoders costs $C_E \approx 0.090$ TFLOPs, and its backward pass costs approximately the same, because the encoders are frozen and only input gradients are needed. Each training sample therefore costs approximately $3K\cdot C_G + 2K\cdot C_E = 1.572$ TFLOPs for $K=4$. Training for 256K iterations with a batch size of 32 processes 8.192M training samples, for an estimated total cost of 12.88 EFLOPs, which closely matches our measured 13.1 EFLOPs.

This is equivalent to 230K PRAFAttack runs with 3 surrogate encoders (57.0 TFLOPs per attack) or 14.7K runs with 5 surrogate encoders (893.3 TFLOPs per attack). As shown in Appendix \ref{app:intermediate_step_results_amortized_praf5}, for the 5-encoder configuration, 100 steps of PRAFAttack optimization with SPARK's amortized initialization is comparable to 300 steps with random initialization. The breakeven point is therefore $N^* = C_{\text{train}} / (C_{300} - C_{100} - C_{\text{SPARK}}) = 13.1\text{ EFLOPs} / (893.3 - 297.8 - 0.37)\text{ TFLOPs} \approx 22$K images, where $C_T$ is the cost of a $T$-step attack and $C_{\text{SPARK}}$ is the cost of generating the initialization.

\subsection{Training and Validation Loss Curves}
\label{app:loss-curves}

\begin{figure}[h]
    \centering
    \includegraphics[width=1\linewidth]{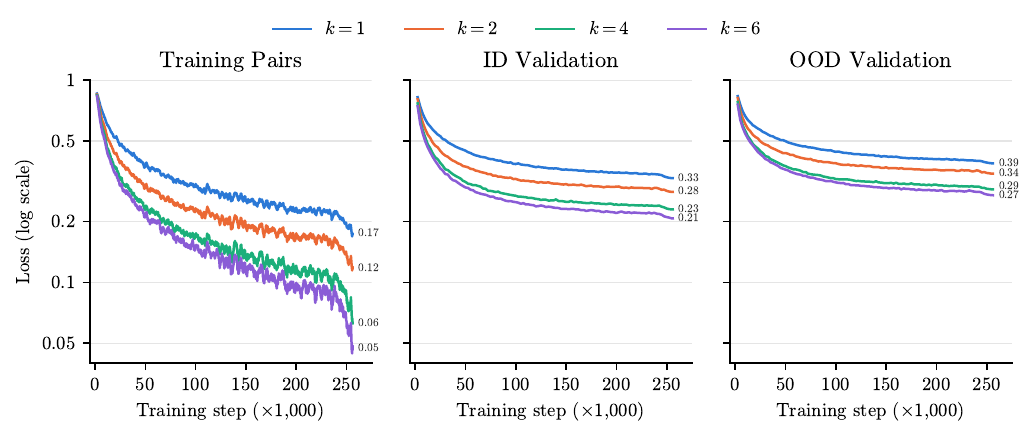}
    \caption{Final-call PRAF loss during training for SPARK with tied weights at different values of K.}
    \label{fig:training-loss-k}
\end{figure}

\clearpage
\section{Additional Results}
\label{app:detailed-results}

\subsection{ASR vs. Optimization Step for OOD-test and ObjectNet-test}
\label{app:asr-vs-steps-ood-objectnet}

\begin{figure}[h]
    \centering
    \begin{subfigure}[t]{0.7\linewidth}
        \centering
        \includegraphics[width=\linewidth]{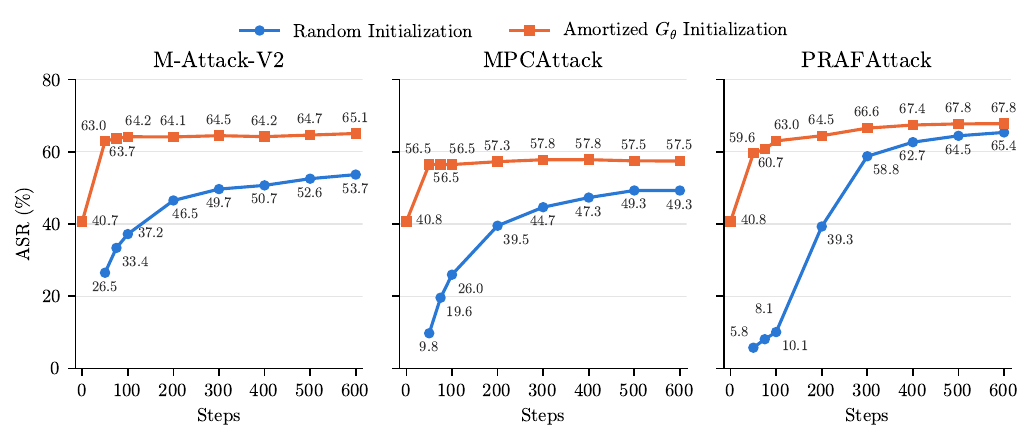}
        \caption{\texttt{OOD-test}.}
        \label{fig:asr-vs-steps-ood}
    \end{subfigure}

    \vspace{0.5em}

    \begin{subfigure}[t]{0.7\linewidth}
        \centering
        \includegraphics[width=\linewidth]{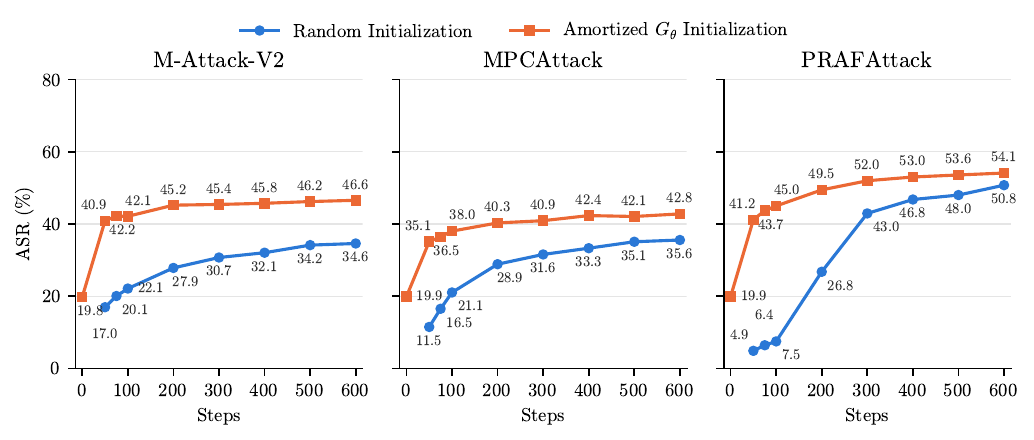}
        \caption{\texttt{Objectnet-test}.}
        \label{fig:asr-vs-steps-objectnet}
    \end{subfigure}

    \caption{Average ASR (\%) vs. attack steps for random and amortized initialization (K=4 SPARK) of gradient-based attacks on (a) \texttt{OOD-test} and (b) \texttt{Objectnet-test}. ASR is averaged over 6 open-source black-box VLMs. All attacks use the default 3-encoder surrogate ensemble.}
    \label{fig:asr-vs-steps-ood-objectnet}
\end{figure}

\subsection{ASR vs. Optimization Step for PRAFAttack with 5-encoder surrogate ensemble}
\label{app:intermediate_step_results_amortized_praf5}
    
\begin{figure}
    \centering
    \includegraphics[width=0.7\linewidth]{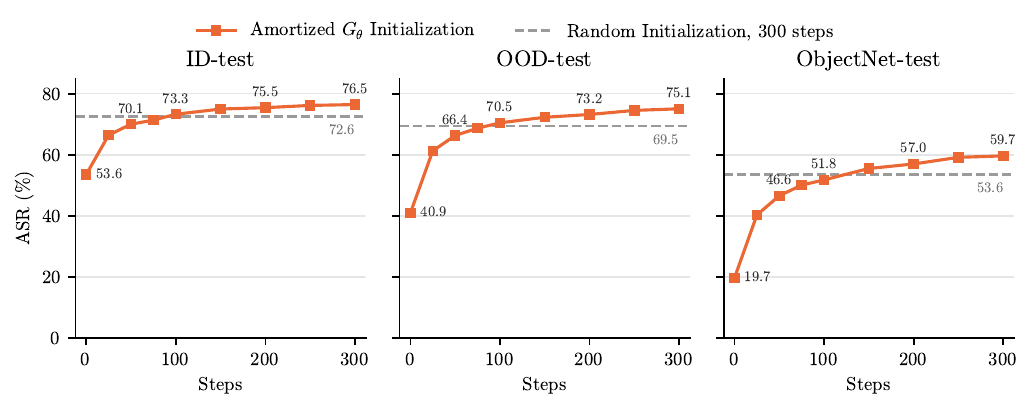}
    \caption{Average ASR (\%) vs. attack steps for amortized initialization (K=4 SPARK) of PRAFAttack using 5 surrogate encoders: OpenAI CLIP ViT-B/16 and ViT-B/32, LAION CLIP ViT-g/14, DINOv2-base, and InternVL3-1B vision encoder. ASR is averaged over 6 open-source black-box VLMs.}
    \label{fig:intermediate_step_results_amortized_praf5}
\end{figure}

\clearpage
\subsection{ASR at Different GPTScore Thresholds}
\label{app:gptscore-thresholds}

\begin{figure}[h]
    \centering
    
    \begin{subfigure}[t]{\linewidth}
        \centering
        \includegraphics[width=\linewidth]{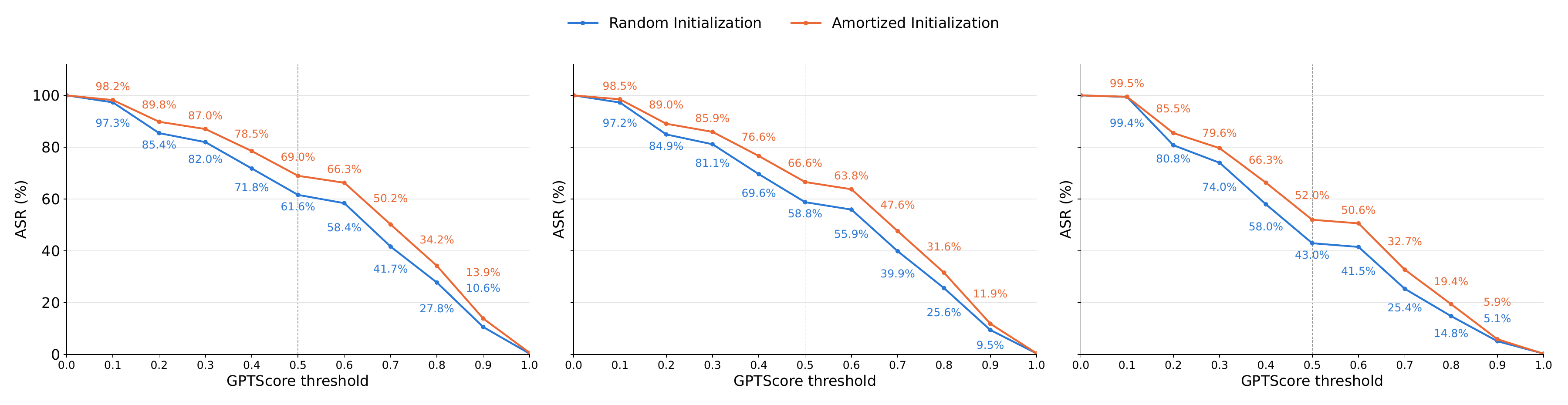}
        \caption{PRAFAttack at 300 steps.}
    \end{subfigure}

    \begin{subfigure}[t]{\linewidth}
        \centering
        \includegraphics[width=\linewidth]{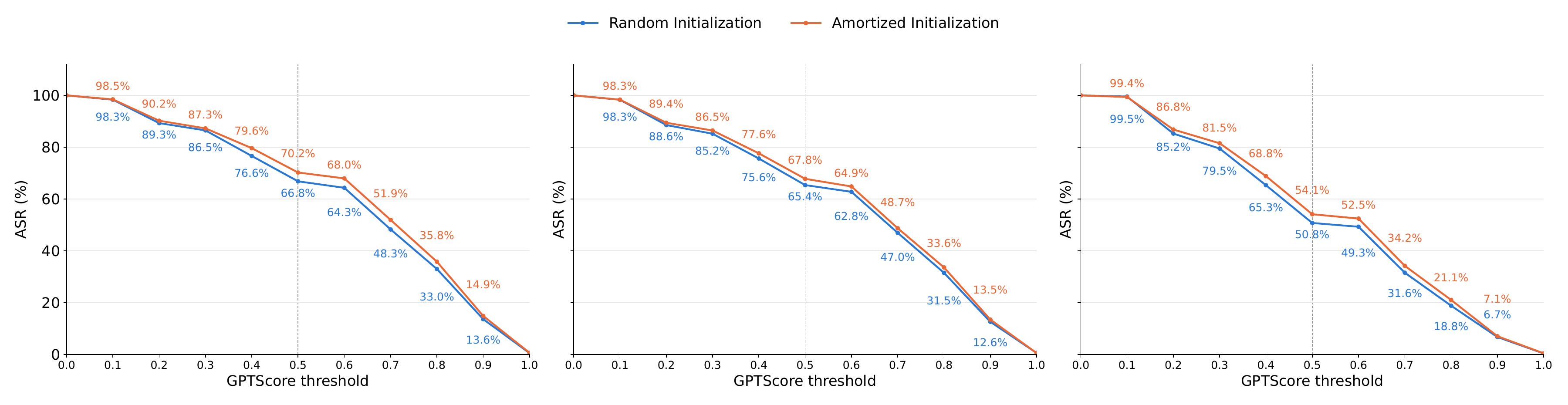}
        \caption{PRAFAttack at 600 steps.}
    \end{subfigure}

\caption{Average ASR (\%) over 6 open-source VLMs at different GPTScore thresholds for PRAFAttack with the shared 3 surrogate ensemble. ASR is computed as the fraction of samples where GPTScore is greater than the corresponding threshold. Amortized initialization maintains a consistent advantage at almost all thresholds.}
\label{fig:asr-vs-threshold}

\end{figure}

\clearpage
\subsection{Paired Significance Tests for ASR and GPTScore}
\label{app:statistical-significance}

We test the statistical significance of PRAFAttack versus SPARK + PRAFAttack separately for each VLM, combining the three 1,000-pair test sets into 3,000 paired samples. We use a two-sided exact McNemar test for ASR and a two-sided paired $t$-test for GPTScore. Holm correction is applied jointly across all 42 tests in the two tables. Tables \ref{tab:paired-significance-asr} and \ref{tab:paired-significance-gptscore} show that almost all improvements remain statistically significant after correction, except for InternVL3.5 8B and Phi4 RV 15B in the 600-step configuration. All observed differences remain positive.

\begin{table*}[h]
\centering
\small
\caption{Paired \textbf{ASR significance tests} of amortized initialization. PRAFAttack uses 5 surrogate ensembles. Bold $p$-values indicate significance at 0.05.}
\label{tab:paired-significance-asr}
\resizebox{0.8\textwidth}{!}{%
\begin{tabular}{llrrrrr}
\toprule
Comparison & VLM & PRAFAttack & \shortstack{PRAFAttack\\+ SPARK} & Difference [95\% CI] & $p$ & $p_{\mathrm{Holm}}$ \\
\midrule
\multirow[c]{6}{*}{\shortstack{PRAFAttack\\300 steps\\3 surrogates}} & Gemma4 12B & 59.87 & 66.57 & +6.70 [4.90, 8.50] & $\mathbf{5.04\times10^{-13}}$ & $\mathbf{1.21\times10^{-11}}$ \\
 & GLM4.6V Flash & 56.17 & 65.37 & +9.20 [7.44, 10.96] & $\mathbf{3.50\times10^{-24}}$ & $\mathbf{1.19\times10^{-22}}$ \\
 & InternVL3.5 8B & 56.73 & 64.43 & +7.70 [5.97, 9.43] & $\mathbf{5.62\times10^{-18}}$ & $\mathbf{1.74\times10^{-16}}$ \\
 & KimiVL A3B & 52.77 & 60.90 & +8.13 [6.37, 9.89] & $\mathbf{2.77\times10^{-19}}$ & $\mathbf{8.85\times10^{-18}}$ \\
 & Phi4 RV 15B & 38.53 & 45.73 & +7.20 [5.45, 8.95] & $\mathbf{1.20\times10^{-15}}$ & $\mathbf{3.23\times10^{-14}}$ \\
 & Qwen3VL 8B & 62.60 & 72.07 & +9.47 [7.76, 11.17] & $\mathbf{4.05\times10^{-27}}$ & $\mathbf{1.46\times10^{-25}}$ \\
\midrule
\multirow[c]{6}{*}{\shortstack{PRAFAttack\\600 steps\\3 surrogates}} & Gemma4 12B & 66.53 & 70.43 & +3.90 [2.18, 5.62] & $\mathbf{1.03\times10^{-5}}$ & $\mathbf{8.27\times10^{-5}}$ \\
 & GLM4.6V Flash & 63.50 & 66.13 & +2.63 [0.99, 4.28] & $\mathbf{0.0020}$ & $\mathbf{0.0099}$ \\
 & InternVL3.5 8B & 63.57 & 65.30 & +1.73 [0.08, 3.39] & $\mathbf{0.0444}$ & $0.0888$ \\
 & KimiVL A3B & 57.20 & 61.23 & +4.03 [2.37, 5.70] & $\mathbf{2.58\times10^{-6}}$ & $\mathbf{2.58\times10^{-5}}$ \\
 & Phi4 RV 15B & 45.17 & 47.17 & +2.00 [0.26, 3.74] & $\mathbf{0.0270}$ & $0.0809$ \\
 & Qwen3VL 8B & 70.00 & 74.07 & +4.07 [2.49, 5.65] & $\mathbf{5.78\times10^{-7}}$ & $\mathbf{7.51\times10^{-6}}$ \\
\midrule
\multirow[c]{9}{*}{\shortstack{PRAFAttack\\300 steps\\5 surrogates}} & Gemma4 12B & 72.87 & 78.07 & +5.20 [3.54, 6.86] & $\mathbf{1.07\times10^{-9}}$ & $\mathbf{1.93\times10^{-8}}$ \\
 & GLM4.6V Flash & 65.43 & 70.90 & +5.47 [3.80, 7.13] & $\mathbf{1.59\times10^{-10}}$ & $\mathbf{3.33\times10^{-9}}$ \\
 & InternVL3.5 8B & 71.60 & 75.80 & +4.20 [2.59, 5.81] & $\mathbf{3.81\times10^{-7}}$ & $\mathbf{5.71\times10^{-6}}$ \\
 & KimiVL A3B & 61.53 & 66.47 & +4.93 [3.27, 6.60] & $\mathbf{7.83\times10^{-9}}$ & $\mathbf{1.33\times10^{-7}}$ \\
 & Phi4 RV 15B & 46.67 & 49.20 & +2.53 [0.78, 4.29] & $\mathbf{0.0053}$ & $\mathbf{0.0211}$ \\
 & Qwen3VL 8B & 73.47 & 79.83 & +6.37 [4.82, 7.91] & $\mathbf{8.90\times10^{-16}}$ & $\mathbf{2.49\times10^{-14}}$ \\
 & GPT-6.1 Sol & 26.23 & 30.57 & +4.33 [2.66, 6.01] & $\mathbf{5.13\times10^{-7}}$ & $\mathbf{7.18\times10^{-6}}$ \\
 & Claude Opus 5.5 & 59.90 & 63.90 & +4.00 [2.31, 5.69] & $\mathbf{4.18\times10^{-6}}$ & $\mathbf{3.76\times10^{-5}}$ \\
 & Gemini 4 Argon & 38.47 & 44.60 & +6.13 [4.23, 8.03] & $\mathbf{3.42\times10^{-10}}$ & $\mathbf{6.84\times10^{-9}}$ \\
\bottomrule
\end{tabular}%
}
\end{table*}

\begin{table*}[h]
\centering
\small
\caption{Paired \textbf{GPTScore significance tests} of amortized initialization. PRAFAttack uses 5 surrogate ensembles. Bold $p$-values indicate significance at 0.05.}
\label{tab:paired-significance-gptscore}
\resizebox{0.8\textwidth}{!}{%
\begin{tabular}{llrrrrr}
\toprule
Comparison & VLM & PRAFAttack & \shortstack{PRAFAttack\\+ SPARK} & Difference [95\% CI] & $p$ & $p_{\mathrm{Holm}}$ \\
\midrule
\multirow[c]{6}{*}{\shortstack{PRAFAttack\\300 steps\\3 surrogates}} & Gemma4 12B & 0.5407 & 0.5847 & +0.0440 [0.0361, 0.0518] & $\mathbf{1.20\times10^{-27}}$ & $\mathbf{4.44\times10^{-26}}$ \\
 & GLM4.6V Flash & 0.5092 & 0.5643 & +0.0550 [0.0470, 0.0630] & $\mathbf{2.07\times10^{-40}}$ & $\mathbf{8.27\times10^{-39}}$ \\
 & InternVL3.5 8B & 0.5211 & 0.5645 & +0.0434 [0.0357, 0.0511] & $\mathbf{5.81\times10^{-28}}$ & $\mathbf{2.21\times10^{-26}}$ \\
 & KimiVL A3B & 0.4834 & 0.5352 & +0.0519 [0.0436, 0.0601] & $\mathbf{4.33\times10^{-34}}$ & $\mathbf{1.69\times10^{-32}}$ \\
 & Phi4 RV 15B & 0.3946 & 0.4395 & +0.0449 [0.0364, 0.0535] & $\mathbf{2.10\times10^{-24}}$ & $\mathbf{7.35\times10^{-23}}$ \\
 & Qwen3VL 8B & 0.5565 & 0.6168 & +0.0603 [0.0525, 0.0681] & $\mathbf{7.22\times10^{-50}}$ & $\mathbf{3.03\times10^{-48}}$ \\
\midrule
\multirow[c]{6}{*}{\shortstack{PRAFAttack\\600 steps\\3 surrogates}} & Gemma4 12B & 0.5820 & 0.6006 & +0.0186 [0.0113, 0.0259] & $\mathbf{6.60\times10^{-7}}$ & $\mathbf{7.92\times10^{-6}}$ \\
 & GLM4.6V Flash & 0.5554 & 0.5755 & +0.0202 [0.0127, 0.0276] & $\mathbf{1.10\times10^{-7}}$ & $\mathbf{1.76\times10^{-6}}$ \\
 & InternVL3.5 8B & 0.5597 & 0.5752 & +0.0154 [0.0080, 0.0228] & $\mathbf{4.19\times10^{-5}}$ & $\mathbf{2.93\times10^{-4}}$ \\
 & KimiVL A3B & 0.5214 & 0.5408 & +0.0194 [0.0116, 0.0271] & $\mathbf{9.67\times10^{-7}}$ & $\mathbf{1.06\times10^{-5}}$ \\
 & Phi4 RV 15B & 0.4415 & 0.4491 & +0.0076 [-0.0008, 0.0160] & $0.0767$ & $0.0888$ \\
 & Qwen3VL 8B & 0.6064 & 0.6290 & +0.0226 [0.0154, 0.0297] & $\mathbf{6.67\times10^{-10}}$ & $\mathbf{1.27\times10^{-8}}$ \\
\midrule
\multirow[c]{9}{*}{\shortstack{PRAFAttack\\300 steps\\5 surrogates}} & Gemma4 12B & 0.6209 & 0.6514 & +0.0305 [0.0232, 0.0379] & $\mathbf{4.93\times10^{-16}}$ & $\mathbf{1.43\times10^{-14}}$ \\
 & GLM4.6V Flash & 0.5682 & 0.6036 & +0.0354 [0.0279, 0.0430] & $\mathbf{9.06\times10^{-20}}$ & $\mathbf{2.99\times10^{-18}}$ \\
 & InternVL3.5 8B & 0.6186 & 0.6448 & +0.0261 [0.0189, 0.0334] & $\mathbf{1.87\times10^{-12}}$ & $\mathbf{4.30\times10^{-11}}$ \\
 & KimiVL A3B & 0.5424 & 0.5740 & +0.0316 [0.0237, 0.0395] & $\mathbf{6.40\times10^{-15}}$ & $\mathbf{1.66\times10^{-13}}$ \\
 & Phi4 RV 15B & 0.4504 & 0.4651 & +0.0147 [0.0061, 0.0233] & $\mathbf{8.09\times10^{-4}}$ & $\mathbf{0.0049}$ \\
 & Qwen3VL 8B & 0.6256 & 0.6742 & +0.0485 [0.0415, 0.0555] & $\mathbf{1.13\times10^{-40}}$ & $\mathbf{4.63\times10^{-39}}$ \\
 & GPT-6.1 Sol & 0.3132 & 0.3420 & +0.0288 [0.0203, 0.0373] & $\mathbf{3.62\times10^{-11}}$ & $\mathbf{7.97\times10^{-10}}$ \\
 & Claude Opus 5.5 & 0.5351 & 0.5623 & +0.0272 [0.0199, 0.0344] & $\mathbf{2.62\times10^{-13}}$ & $\mathbf{6.54\times10^{-12}}$ \\
 & Gemini 4 Argon & 0.4013 & 0.4406 & +0.0393 [0.0304, 0.0481] & $\mathbf{6.28\times10^{-18}}$ & $\mathbf{1.88\times10^{-16}}$ \\
\bottomrule
\end{tabular}%
}
\end{table*}

\clearpage
\subsection{Evaluation Prompts and Judge Sensitivity}
\label{app:gptscore-judge}

For every attacked image and its corresponding target image, the evaluated VLM generates a short description using the following prompt.

\noindent
\begin{tcolorbox}[
    enhanced,
    title={Concise Description Prompt},
    width=\textwidth,
    colback=black!6,
    colframe=black!6,
    colbacktitle=black!15,
    coltitle=black,
    fonttitle=\bfseries,
    boxrule=0pt,
    toptitle=3pt,
    bottomtitle=3pt,
    arc=2pt,
    outer arc=2pt,
    boxsep=0pt,
    left=10pt,
    right=10pt,
    top=9pt,
    bottom=9pt,
    before skip=8pt,
    after skip=10pt
]
\small
Describe this image in one concise sentence, no longer than 20 words.
\end{tcolorbox}

We compare the semantic similarity between the two textual descriptions using GPTScore \citep{fu2024gptscore} with the following prompt.

\noindent
\begin{tcolorbox}[
    enhanced,
    title={GPTScore Prompt},
    width=\textwidth,
    colback=black!6,
    colframe=black!6,
    colbacktitle=black!15,
    coltitle=black,
    fonttitle=\bfseries,
    boxrule=0pt,
    toptitle=3pt,
    bottomtitle=3pt,
    arc=2pt,
    outer arc=2pt,
    boxsep=0pt,
    left=10pt,
    right=10pt,
    top=9pt,
    bottom=9pt,
    before skip=8pt,
    after skip=10pt
]
\small
Rate the semantic similarity between the following two texts on a scale from 0 to 1.

\medskip
\textbf{Criteria for similarity measurement:}
\begin{enumerate}
    \setlength{\itemsep}{2pt}
    \setlength{\parskip}{0pt}
    \setlength{\topsep}{4pt}
    \item \textbf{Main Subject Consistency:} If both descriptions refer to the same key subject or object (e.g., a person, food, an event), they should receive a higher similarity score.
    \item \textbf{Relevant Description:} If the descriptions are related to the same context or topic, they should also contribute to a higher similarity score.
    \item \textbf{Ignore Fine-Grained Details:} Do not penalize differences in \textbf{phrasing, sentence structure, or minor variations in detail}. Focus on \textbf{whether both descriptions fundamentally describe the same thing.}
    \item \textbf{Partial Matches:} If one description contains extra information but does not contradict the other, they should still have a high similarity score.
    \item \textbf{Similarity Score Range:}
    \begin{itemize}
        \setlength{\itemsep}{1pt}
        \item \textbf{1.0}: Nearly identical in meaning.
        \item \textbf{0.8--0.9}: Same subject, with highly related descriptions.
        \item \textbf{0.7--0.8}: Same subject, core meaning aligned, even if some details differ.
        \item \textbf{0.5--0.7}: Same subject but different perspectives or missing details.
        \item \textbf{0.3--0.5}: Related but not highly similar (same general theme but different descriptions).
        \item \textbf{0.0--0.2}: Completely different subjects or unrelated meanings.
    \end{itemize}
\end{enumerate}

\textbf{Text 1:} \{target-image description\}

\textbf{Text 2:} \{attacked-image description\}

\medskip
Output only a single number between 0 and 1. Do not include any explanation or additional text.
\end{tcolorbox}

\citet{prafattack} binarizes the score as $\mathrm{AttackSuccess}=\mathbf{1}\{\mathrm{GPTScore}\geq0.5\}$. We use \texttt{Qwen/Qwen3.8-27B} as the GPTScore scoring model for all experiments reported in the main manuscript and in the appendix.

\begin{figure}[h]
    \centering
    \includegraphics[width=0.7\linewidth]{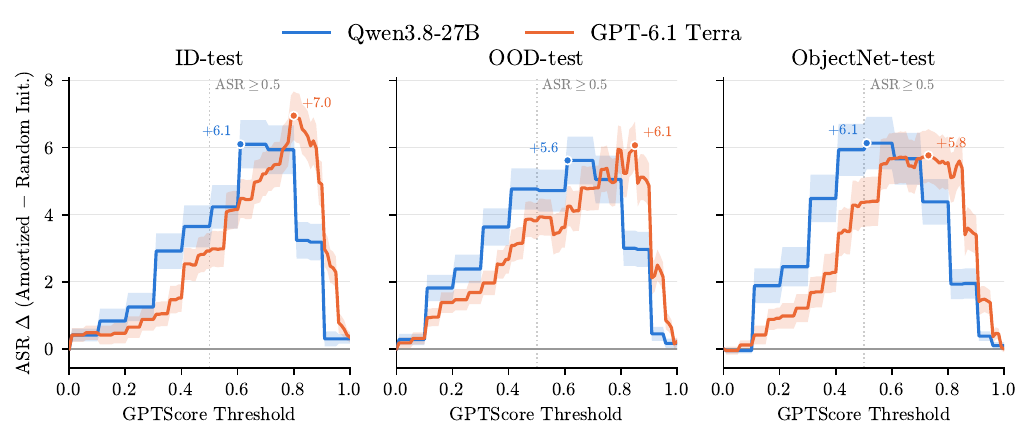}
    \caption{\textbf{ASR gain across judge thresholds.} Difference in ASR between amortized
    (PRAFAttack + SPARK) and random initialization as the score required for success varies, under both
    judges; bands are $\pm 1$ SEM.}
    \label{fig:judge-at-thresholds}
\end{figure}

\clearpage
\textbf{Agreement and ASR Across Judges.} We test the effect of using a different scoring model, \texttt{GPT-6.1 Terra}. The two judges' scores agree closely across all 36{,}000 caption pairs (Pearson $r = 0.937$), with GPT-6.1 Terra scoring $0.116$ higher on average. Table~\ref{tab:judge-sensitivity} shows that at the fixed $0.5$ threshold, this leniency affects the ASR improvement from amortized initialization: the gain in mean score is the same under both judges, but the more lenient judge already counts most pairs as successful at $0.5$, leaving fewer failures to convert. Figure~\ref{fig:judge-at-thresholds} shows the average $\Delta$ ASR across thresholds for both judges.

\providecommand{\dcell}[1]{{\footnotesize\textit{#1}}}
\begin{table*}[h]
    \centering
    \renewcommand{\arraystretch}{1.2}
    \setlength{\aboverulesep}{0pt}\setlength{\belowrulesep}{0pt}\setlength{\extrarowheight}{1pt}\setlength{\tabcolsep}{3pt}
    \caption{\textbf{Judge sensitivity.} Per-victim ASR (score $\geq 0.5$) and mean judge score for
    PRAFAttack and PRAFAttack + SPARK, with the same attacked images and victim captions scored by two
    judges, Qwen-3.8-27B and GPT 6.1 Terra.}
    \label{tab:judge-sensitivity}
    \resizebox{\textwidth}{!}{%
    \begin{tabular}{l l cc cc cc cc cc cc | cc}
    \toprule
     & & \multicolumn{2}{c}{Gemma-4-12B} & \multicolumn{2}{c}{Qwen3-VL-8B} & \multicolumn{2}{c}{InternVL3.5-8B} & \multicolumn{2}{c}{GLM-4.6V-Flash} & \multicolumn{2}{c}{Kimi-VL-A3B} & \multicolumn{2}{c|}{Phi-4-RV-15B} & \multicolumn{2}{c}{Avg.} \\
    \cmidrule(lr){3-4} \cmidrule(lr){5-6} \cmidrule(lr){7-8} \cmidrule(lr){9-10} \cmidrule(lr){11-12} \cmidrule(lr){13-14} \cmidrule(lr){15-16}
    Judge & Method & ASR & AvgScore & ASR & AvgScore & ASR & AvgScore & ASR & AvgScore & ASR & AvgScore & ASR & AvgScore & ASR & AvgScore \\
    \midrule
    \multicolumn{16}{c}{\texttt{ID-test-S}} \\
    \midrule
    \rowcolor{blue!5}  & PRAFAttack & 77.9 & 0.651 & 79.4 & 0.660 & 79.2 & 0.666 & 74.0 & 0.614 & 70.0 & 0.586 & 55.1 & 0.492 & 72.6 & 0.612 \\
    \rowcolor{blue!5} Qwen3.8-27B & \textcolor{magenta}{$+$ SPARK} & \textbf{83.4} & \textbf{0.682} & \textbf{84.0} & \textbf{0.708} & \textbf{82.2} & \textbf{0.687} & \textbf{77.1} & \textbf{0.641} & \textbf{73.5} & \textbf{0.615} & \textbf{57.3} & \textbf{0.510} & \textbf{76.2} & \textbf{0.640} \\
    \rowcolor{blue!15}  & \quad\textit{$\Delta$} & \dcell{+5.5} & \dcell{+0.031} & \dcell{+4.6} & \dcell{+0.047} & \dcell{+3.0} & \dcell{+0.021} & \dcell{+3.1} & \dcell{+0.027} & \dcell{+3.5} & \dcell{+0.028} & \dcell{+2.2} & \dcell{+0.018} & \dcell{+3.7} & \dcell{+0.029} \\
    \cmidrule(lr){1-16}
    \rowcolor{orange!5}  & PRAFAttack & 89.2 & 0.768 & 87.2 & 0.762 & 87.8 & 0.769 & 82.1 & 0.725 & 80.1 & 0.704 & 67.9 & 0.617 & 82.4 & 0.724 \\
    \rowcolor{orange!5} GPT-6.1 Terra & \textcolor{magenta}{$+$ SPARK} & \textbf{91.2} & \textbf{0.795} & \textbf{91.8} & \textbf{0.803} & \textbf{89.1} & \textbf{0.791} & \textbf{85.3} & \textbf{0.752} & \textbf{83.5} & \textbf{0.733} & \textbf{70.9} & \textbf{0.638} & \textbf{85.3} & \textbf{0.752} \\
    \rowcolor{orange!15}  & \quad\textit{$\Delta$} & \dcell{+2.0} & \dcell{+0.027} & \dcell{+4.6} & \dcell{+0.041} & \dcell{+1.3} & \dcell{+0.021} & \dcell{+3.2} & \dcell{+0.027} & \dcell{+3.4} & \dcell{+0.029} & \dcell{+3.0} & \dcell{+0.020} & \dcell{+2.9} & \dcell{+0.028} \\
    \midrule
    \multicolumn{16}{c}{\texttt{OOD-test-S}} \\
    \midrule
    \rowcolor{blue!5}  & PRAFAttack & 77.8 & 0.646 & 76.6 & 0.645 & 77.0 & 0.650 & 69.6 & 0.592 & 65.4 & 0.564 & 50.8 & 0.472 & 69.5 & 0.595 \\
    \rowcolor{blue!5} Qwen3.8-27B & \textcolor{magenta}{$+$ SPARK} & \textbf{80.5} & \textbf{0.668} & \textbf{83.2} & \textbf{0.695} & \textbf{81.5} & \textbf{0.679} & \textbf{75.7} & \textbf{0.629} & \textbf{71.5} & \textbf{0.599} & \textbf{53.4} & \textbf{0.488} & \textbf{74.3} & \textbf{0.626} \\
    \rowcolor{blue!15}  & \quad\textit{$\Delta$} & \dcell{+2.7} & \dcell{+0.022} & \dcell{+6.6} & \dcell{+0.050} & \dcell{+4.5} & \dcell{+0.029} & \dcell{+6.1} & \dcell{+0.037} & \dcell{+6.1} & \dcell{+0.035} & \dcell{+2.6} & \dcell{+0.017} & \dcell{+4.8} & \dcell{+0.032} \\
    \cmidrule(lr){1-16}
    \rowcolor{orange!5}  & PRAFAttack & 88.8 & 0.765 & 85.6 & 0.748 & 86.8 & 0.762 & 79.4 & 0.704 & 77.6 & 0.687 & 63.3 & 0.589 & 80.2 & 0.709 \\
    \rowcolor{orange!5} GPT-6.1 Terra & \textcolor{magenta}{$+$ SPARK} & \textbf{90.3} & \textbf{0.782} & \textbf{90.2} & \textbf{0.790} & \textbf{89.3} & \textbf{0.784} & \textbf{84.5} & \textbf{0.740} & \textbf{82.7} & \textbf{0.718} & \textbf{67.4} & \textbf{0.613} & \textbf{84.1} & \textbf{0.738} \\
    \rowcolor{orange!15}  & \quad\textit{$\Delta$} & \dcell{+1.5} & \dcell{+0.017} & \dcell{+4.6} & \dcell{+0.042} & \dcell{+2.5} & \dcell{+0.022} & \dcell{+5.1} & \dcell{+0.037} & \dcell{+5.1} & \dcell{+0.031} & \dcell{+4.1} & \dcell{+0.024} & \dcell{+3.8} & \dcell{+0.029} \\
    \midrule
    \multicolumn{16}{c}{\texttt{ObjectNet-test}} \\
    \midrule
    \rowcolor{blue!5}  & PRAFAttack & 62.9 & 0.566 & 64.4 & 0.572 & 58.6 & 0.540 & 52.7 & 0.498 & 49.2 & 0.477 & 34.1 & 0.387 & 53.6 & 0.507 \\
    \rowcolor{blue!5} Qwen3.8-27B & \textcolor{magenta}{$+$ SPARK} & \textbf{70.3} & \textbf{0.605} & \textbf{72.3} & \textbf{0.620} & \textbf{63.7} & \textbf{0.568} & \textbf{59.9} & \textbf{0.540} & \textbf{54.4} & \textbf{0.508} & \textbf{36.9} & \textbf{0.397} & \textbf{59.6} & \textbf{0.540} \\
    \rowcolor{blue!15}  & \quad\textit{$\Delta$} & \dcell{+7.4} & \dcell{+0.039} & \dcell{+7.9} & \dcell{+0.048} & \dcell{+5.1} & \dcell{+0.028} & \dcell{+7.2} & \dcell{+0.042} & \dcell{+5.2} & \dcell{+0.031} & \dcell{+2.8} & \dcell{+0.010} & \dcell{+5.9} & \dcell{+0.033} \\
    \cmidrule(lr){1-16}
    \rowcolor{orange!5}  & PRAFAttack & 81.0 & 0.697 & 76.1 & 0.680 & 74.1 & 0.668 & 67.3 & 0.624 & 64.2 & 0.608 & 48.8 & 0.510 & 68.6 & 0.631 \\
    \rowcolor{orange!5} GPT-6.1 Terra & \textcolor{magenta}{$+$ SPARK} & \textbf{83.7} & \textbf{0.726} & \textbf{82.6} & \textbf{0.725} & \textbf{77.8} & \textbf{0.690} & \textbf{74.1} & \textbf{0.665} & \textbf{69.5} & \textbf{0.635} & \textbf{50.0} & \textbf{0.520} & \textbf{73.0} & \textbf{0.660} \\
    \rowcolor{orange!15}  & \quad\textit{$\Delta$} & \dcell{+2.7} & \dcell{+0.030} & \dcell{+6.5} & \dcell{+0.045} & \dcell{+3.7} & \dcell{+0.022} & \dcell{+6.8} & \dcell{+0.041} & \dcell{+5.3} & \dcell{+0.028} & \dcell{+1.2} & \dcell{+0.010} & \dcell{+4.4} & \dcell{+0.029} \\
    \bottomrule
    \end{tabular}}
\end{table*}

\clearpage
\section{SPARK Ablation Studies}
\label{app:ablations}

\subsection{Training Objective and Intermediate Supervision}
\label{app:training-objective-ablation}

We investigate how the choice of training objective and the supervision of intermediate refinement steps affect the transferability of SPARK's generated perturbations. Our default model uses the PRAFAttack-based loss described in Section~\ref{sec:method} and applies supervision after each of the $K=4$ refinement steps.

\noindent\textbf{Intermediate supervision.}
To evaluate the importance of supervising intermediate steps, we train a variant of SPARK using the same architecture and training objective, but apply the loss only to the final perturbation. On \texttt{ID-test}, this reduces standalone ASR from 53.4\% to 52.2\% for $K=4$. The difference becomes larger when evaluating with six refinement steps, where ASR decreases from 54.9\% to 52.5\%. These results suggest that intermediate supervision helps the shared network learn refinement steps that remain effective when applied repeatedly.

\noindent\textbf{Training objective.}
We also replace the PRAFAttack-based training loss with the InfoNCE objective used by MPCAttack, while keeping the generator architecture and all other training hyperparameters unchanged. At $K=4$, ASR on \texttt{ID-test} decreases from 53.4\% to 41.6\%. This reduction indicates that the choice of feature alignment objective plays an important role in learning transferable perturbations. In particular, the combination of global and intermediate feature alignment used by PRAFAttack appears more effective for training SPARK than the contrastive objective considered here.

\begin{table}[h]
\centering
\caption{Training objective ablation.}
\label{tab:training-objective-ablation}
\begin{tabular}{l l cc cc cc}
\toprule
 & & \multicolumn{2}{c}{ID Test} & \multicolumn{2}{c}{OOD Test} & \multicolumn{2}{c}{ObjectNet Test} \\
\cmidrule(lr){3-4} \cmidrule(lr){5-6} \cmidrule(lr){7-8}
Loss & \shortstack{Supervised\\Iteration} & ASR & AvgScore & ASR & AvgScore & ASR & AvgScore \\
\midrule
\multicolumn{8}{l}{SPARK ($K{=}4$)} \\
PRAF & $[1...K]$ & 53.4 & 0.484 & 40.8 & 0.413 & 19.9 & 0.293 \\
PRAF & Only $K^{th}$ & 52.2 & 0.481 & 39.5 & 0.408 & 18.9 & 0.291 \\
InfoNCE & $[1...K]$ & 41.6 & 0.431 & 30.7 & 0.372 & 13.3 & 0.280 \\
\midrule
\multicolumn{8}{l}{SPARK  ($K{=}6$)} \\
PRAF & $[1...K]$ & 54.9 & 0.497 & 42.8 & 0.427 & 20.7 & 0.303 \\
PRAF & Only $K^{th}$ & 52.5 & 0.482 & 39.0 & 0.403 & 17.6 & 0.282 \\
\bottomrule
\end{tabular}
\end{table}

\clearpage
\subsection{Generator Architecture}
\label{app:architecture-ablation}

We evaluate several architectural choices to understand whether SPARK's performance comes from its looped refinement structure, parameter sharing, or simply the computational capacity of the generator. We compare the default model against variants that remove weight sharing, replace iterative refinement with a wider single-pass network, or remove the Transformer bottleneck.

\begin{figure}[h]
    \centering
    \includegraphics[width=0.4\linewidth]{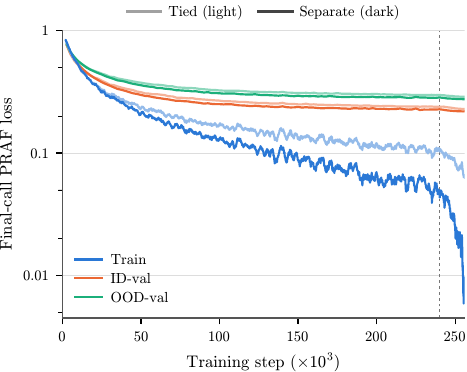}
    \caption{Final-call PRAF loss during training $K=4$ looped generator with and without weight-tying.}
    \label{fig:weight-tying-loss}
\end{figure}

\noindent\textbf{Weight sharing.}
The default SPARK model applies the same network at every refinement step, using 9.8M parameters. We compare this approach with a variant that uses a separate network for each of the four steps, increasing the parameter count to 39.1M. Despite its larger capacity, the unshared model achieves 50.8\% ASR on \texttt{ID-test}, compared with 53.4\% for the shared model. Figure \ref{fig:weight-tying-loss} shows that the unshared model achieves lower training and validation losses but worse black-box transfer. This suggests that parameter sharing may help the model avoid overfitting to the surrogate encoders.

\noindent\textbf{Iterative refinement versus model capacity.}
To determine whether the benefits of looping can be explained by increased computation, we compare SPARK with a wider single-pass generator requiring a similar number of FLOPs. The looped model achieves 53.4\% ASR on \texttt{ID-test}, while the wider single-pass model reaches 43.7\%. This comparison suggests that repeatedly refining a perturbation is more effective than increasing the capacity of a single forward pass at a comparable computational cost.

\noindent\textbf{Transformer bottleneck.}
We also evaluate the importance of the Transformer bottleneck, which allows information exchange across spatial regions in the U-Net. Removing this component reduces single-pass ASR from 42.2\% to 31.2\%, despite only a modest reduction in computational cost. This suggests that spatial information exchange contributes to the effectiveness of the generated perturbations.

\begin{table}[h]
\centering
\caption{Architecture ablations}
\label{tab:architecture-ablation}

\begin{tabular}{l c c cc cc cc}
\toprule
 &  & \multirow{2}{*}{$\displaystyle\frac{\mathrm{GFLOPs}}{\mathrm{image}}$} & \multicolumn{2}{c}{ID Test} & \multicolumn{2}{c}{OOD Test} & \multicolumn{2}{c}{ObjectNet Test} \\
\cmidrule(lr){4-5} \cmidrule(lr){6-7} \cmidrule(lr){8-9}
Method & Params & & ASR & AvgScore & ASR & AvgScore & ASR & AvgScore \\
\midrule
\multicolumn{9}{l}{\textit{$\mathrm{G_\theta}$ ($K{=}1$)}} \\
Default & 9.8M & 161.0 & 42.2 & 0.415 & 31.6 & 0.347 & 14.2 & 0.251 \\
w/o Transformer & 6.4M & 153.5 & 31.2 & 0.357 & 19.8 & 0.282 & 8.6 & 0.215 \\
2$\times$ Channel Width & 35.5M & 366.9 & 43.7 & 0.423 & 31.8 & 0.351 & 13.7 & 0.250 \\
\midrule
\multicolumn{9}{l}{\textit{$\mathrm{G_\theta}$ ($K{=}4$)}} \\
Tied Weights & 9.8M & 373.1 & \textbf{53.4} & \textbf{0.484} & \textbf{40.8} & \textbf{0.413} & \textbf{19.9} & \textbf{0.293} \\
Separate Weights & 39.1M & 373.1 & \underline{50.8} & \underline{0.468} & \underline{38.9} & \underline{0.403} & \underline{18.7} & \underline{0.286} \\
\bottomrule
\end{tabular}
\end{table}

\clearpage
\subsection{Performance at a Smaller Perturbation Budget}
\label{app:perturbation-budget-ablation}

Our main experiments use an $\ell_\infty$ perturbation budget of $\epsilon=16/255$. Here, we investigate whether a generator trained with this budget can be adapted to the smaller budget $\epsilon=8/255$ without retraining, and whether SPARK initialization remains beneficial when the generator is trained directly with the smaller budget.

\noindent\textbf{Adapting $\epsilon=16/255$ SPARK to $\epsilon=8/255$.}
We consider two simple ways to adapt perturbations generated at $\epsilon=16/255$ to the smaller budget: rescaling the perturbation and clipping it to the new constraint. Both approaches perform poorly, achieving at most 5.3\% ASR on \texttt{ID-test}. This suggests that directly modifying the magnitude of a learned perturbation does not preserve its effectiveness at a substantially smaller budget.

\noindent\textbf{Training SPARK with $\epsilon=8/255$.}
We next train SPARK directly with $\epsilon=8/255$ and $K=4$ looped iterations, leaving all other training hyperparameters the same. The resulting generator achieves 22.1\% standalone ASR on \texttt{ID-test}, comparable to the 22.5\% achieved by PRAFAttack under the same perturbation budget. Using the newly trained generator to initialize PRAFAttack further increases ASR to 25.4\%. These findings suggest that SPARK needs to learn perturbations for the intended budget, but that amortized initialization can still improve gradient-based attacks under tighter perturbation constraints.



\begin{table}[h]
    \centering
    \caption{Noise budget ablations. $\mathrm{G_\theta}$ uses $K{=}4$; PRAFAttack runs the stated number of optimization steps.}
    \label{tab:perturbation-budget-ablation}
    
    \begin{tabular}{l cc cc cc cc}
    \toprule
     & \multicolumn{2}{c}{$\ell_\infty$ ($\epsilon/255$)} & \multicolumn{2}{c}{ID Test} & \multicolumn{2}{c}{OOD Test} & \multicolumn{2}{c}{ObjectNet Test} \\
    \cmidrule(lr){2-3} \cmidrule(lr){4-5} \cmidrule(lr){6-7} \cmidrule(lr){8-9}
    Method & Trained & Evaluated & ASR & AvgScore & ASR & AvgScore & ASR & AvgScore \\
    \midrule
    \multicolumn{9}{l}{\textit{Generative Attack ($K{=}4$)}} \\
    $K^{\mathrm{th}}$ iteration rescaled & 16 & 8 & 2.6 & 0.084 & 1.4 & 0.075 & 3.6 & 0.149 \\
    $K^{\mathrm{th}}$ iteration clipped & 16 & 8 & 3.8 & 0.096 & 2.4 & 0.087 & 3.9 & 0.157 \\
    $[0..K]$ iterations clipped & 16 & 8 & 5.3 & 0.116 & 3.6 & 0.103 & 4.6 & 0.163 \\
    Trained with $\epsilon=8$ & 8 & 8 & 22.1 & 0.279 & 15.7 & 0.232 & 8.9 & 0.210 \\
    \midrule
    \multicolumn{9}{l}{\textit{Gradient-based Attack}} \\
    PRAFAttack (300 steps) & -- & 8 & 22.5 & 0.263 & 19.5 & 0.246 & 19.3 & 0.286 \\
    + $\mathrm{\textcolor{magenta}{SPARK}}_{\epsilon=8}$ & 8 & 8 & 25.4 & 0.287 & 22.6 & 0.271 & 22.0 & 0.304 \\
    \rowcolor{blue!13} \quad\textit{$\Delta$} & & & \dcell{+2.9} & \dcell{+0.023} & \dcell{+3.1} & \dcell{+0.025} & \dcell{+2.7} & \dcell{+0.018} \\
    
    PRAFAttack (600 steps) & -- & 8 & 25.0 & 0.283 & 22.6 & 0.272 & 22.6 & 0.308 \\
    + $\mathrm{\textcolor{magenta}{SPARK}}_{\epsilon=8}$ & 8 & 8 & 26.8 & 0.295 & 23.3 & 0.277 & 23.8 & 0.314 \\
    \rowcolor{blue!13} \quad\textit{$\Delta$} & & & \dcell{+1.7} & \dcell{+0.012} & \dcell{+0.8} & \dcell{+0.004} & \dcell{+1.2} & \dcell{+0.006} \\
    \bottomrule
    \end{tabular}
    \end{table}

Table~\ref{tab:perturbation-budget-ablation} shows the effect of reducing the $\ell_\infty$ budget from $16/255$ to $8/255$.
To evaluate $G_\theta$ trained with $\epsilon = 16/255$ under a smaller budget $\epsilon' = 8/255$ without retraining, we compare three strategies.
\textit{$K^{\mathrm{th}}$ iteration rescaled} scales the final perturbation, $x_{adv} = x_{src} + (\epsilon'/\epsilon)\,\hat{\delta}_K$.
\textit{$K^{\mathrm{th}}$ iteration clipped} runs all $K$ steps unchanged and clips only the final perturbation, $x_{adv} = x_{src} + \mathrm{clip}(\hat{\delta}_K, -\epsilon', \epsilon')$.
\textit{$[0..K]$ iterations clipped} clips $\hat{\delta}_0$ and every $\hat{\delta}_k$ to $[-\epsilon', \epsilon']$, so each step $k$ refines the clipped $\hat{\delta}_{k-1}$.

\clearpage
\subsection{Inference-Time Unrolling Beyond the Training Depth}
\label{sec:inference-depth-unrolling}

Since SPARK shares parameters across steps, a generator trained with $T=4$ refinement steps can be unrolled for $K > T$ steps at inference without retraining. Because $G_\theta$ is conditioned on the step index via a sinusoidal embedding, we map each inference step $j \in \{0, \dots, K-1\}$ to an index $t_j$ within the training range using one of two strategies: \textit{Clamp}, $t_j = \min(j, T-1)$, which repeats the final-step embedding ($0, 1, 2, 3, 3, \dots$); and \textit{Interpolate}, $t_j = j(T-1)/(K-1)$, which spaces the steps uniformly over $[0, T-1]$. Both reduce to the trained generator at $K=T$. The $K=4$ results differ slightly from those in Table~\ref{tab:main-table} since victim VLM responses are generated under their recommended decoding parameters with non-zero temperature.

As shown in Table~\ref{tab:inference-depth-unrolling}, additional inference steps yield modest gains. Unrolling with interpolation improves ASR matches or exceeds the $K=4$ baseline in every setting. Clamping yields similar gains on ID-test and OOD-test but is less consistent across $K$. Improvements saturate around $K=16$ and remain smaller than those from training with more steps, indicating that inference-time unrolling is not a substitute for training at a greater depth.

\begin{table}[h]
    \centering
    \caption{Inference-time unrolling of a SPARK generator trained with $K_{train}=4$ steps, using clamped or interpolated step indices for $K_{inference}>K_{train}$.}
    \label{tab:inference-depth-unrolling}
    \begin{tabular}{lccccccc}
    \toprule
    \multirow{2}{*}{Method} & \multirow{2}{*}{$K_{inference}$} & \multicolumn{2}{c}{ID-test} & \multicolumn{2}{c}{OOD-test} & \multicolumn{2}{c}{ObjectNet-test} \\
    \cmidrule(lr){3-4} \cmidrule(lr){5-6} \cmidrule(lr){7-8}
     &  & ASR & GPTScore & ASR & GPTScore & ASR & GPTScore \\
    \midrule
    \multirow{2}{*}{\textit{Native Training}} & 4 & 53.22 & 0.4888 & 40.98 & 0.4184 & 20.00 & 0.2983 \\
     & 6 & 54.93 & 0.4971 & 42.80 & 0.4265 & 20.67 & 0.3032 \\
    \midrule
    \multirow{4}{*}{\textit{Clamp}} & 8 & \underline{53.52} & 0.4900 & 41.25 & 0.4221 & \underline{20.98} & 0.3030 \\
     & 12 & 52.77 & 0.4894 & 41.58 & 0.4237 & \textbf{21.58} & \textbf{0.3057} \\
     & 16 & \underline{53.52} & \underline{0.4919} & \underline{41.93} & \underline{0.4250} & 20.60 & \underline{0.3047} \\
     & 24 & \textbf{54.27} & \textbf{0.4935} & \textbf{42.33} & \textbf{0.4262} & 20.83 & 0.3018 \\
    \midrule
    \multirow{4}{*}{\textit{Interpolate}} & 8 & \underline{54.22} & 0.4941 & 42.13 & 0.4276 & 20.63 & 0.3033 \\
     & 12 & \underline{54.22} & 0.4949 & 42.73 & \textbf{0.4323} & 20.43 & 0.3032 \\
     & 16 & \textbf{54.38} & \underline{0.4960} & \underline{42.85} & \underline{0.4319} & \underline{21.07} & \underline{0.3081} \\
     & 24 & 54.17 & \textbf{0.4965} & \textbf{42.97} & 0.4311 & \textbf{21.17} & \textbf{0.3092} \\
    \bottomrule
    \end{tabular}
    \end{table}

\clearpage
\section{Qualitative Examples}
\label{app:qualitative-examples}
\begin{figure}[h]
    \centering
    \includegraphics[width=0.9\linewidth]{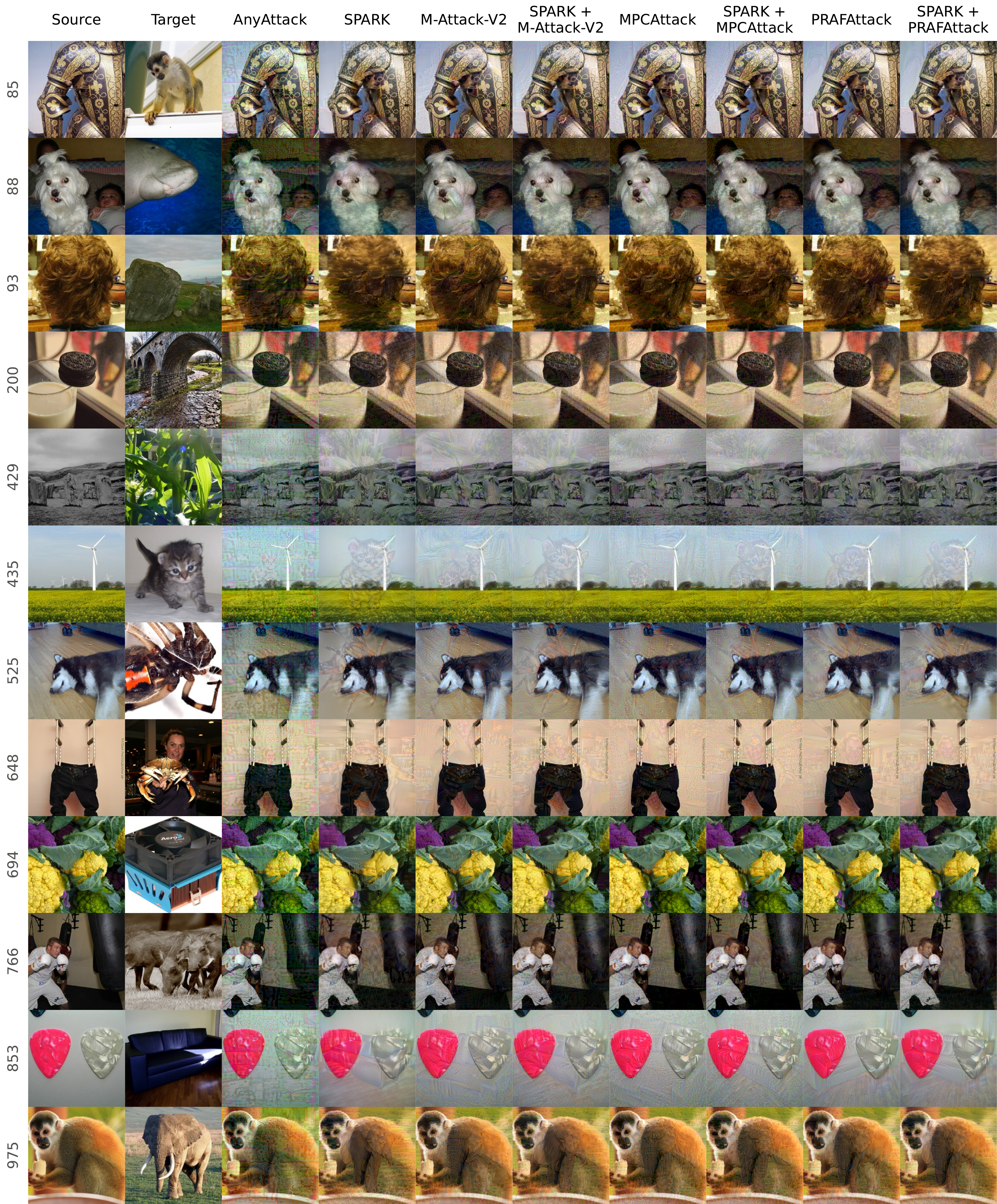}
    \caption{Qualitative examples from \texttt{ObjectNet-test}.}
    \label{fig:qualitative-examples}
\end{figure}

\clearpage
\section{VLM Response Examples}
\label{app:vlm-response-examples}

\begin{figure}[h]
    \centering
    \includegraphics[width=1\linewidth]{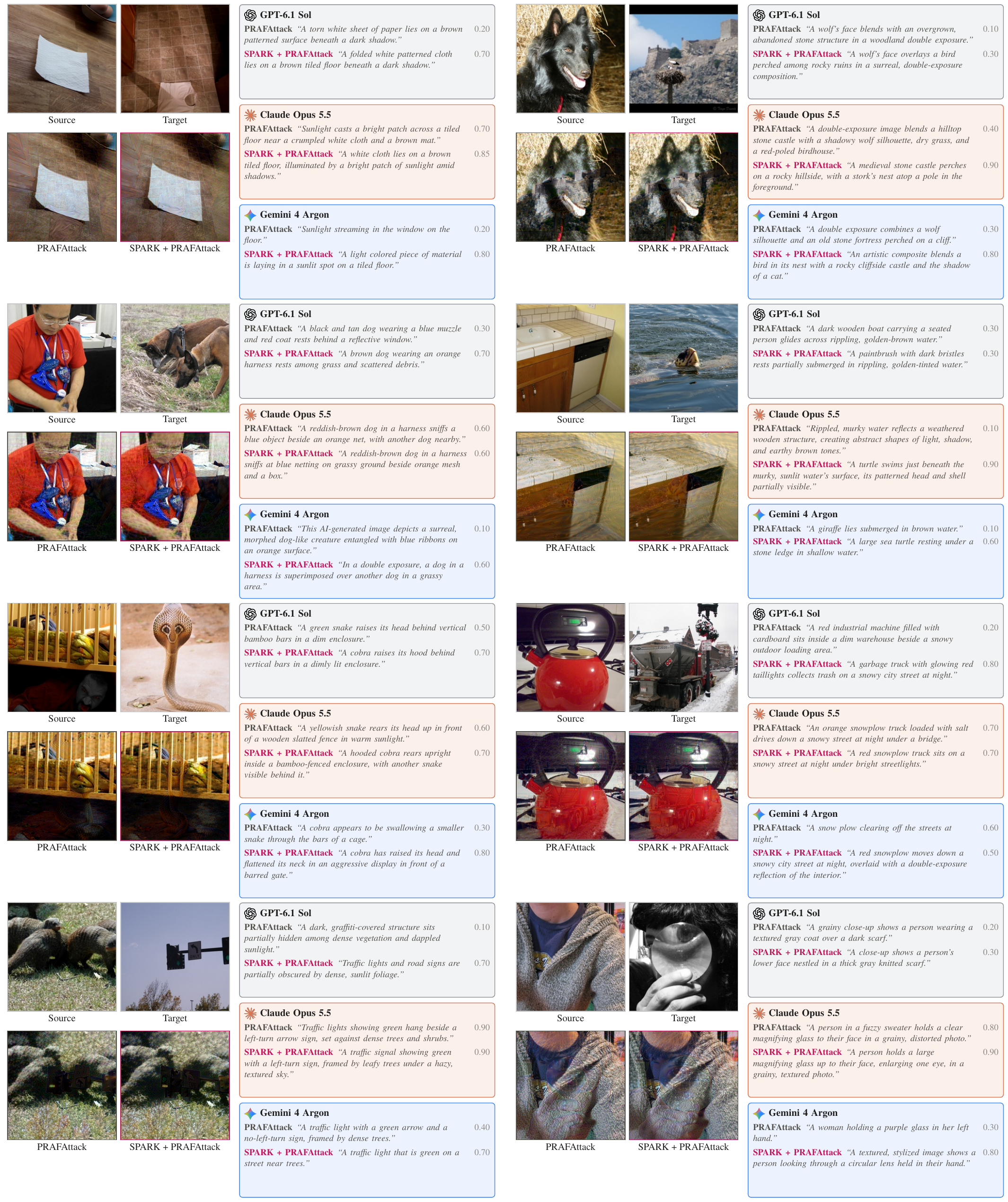}
    \caption{Qualitative VLM response examples for PRAFAttack (5 surrogates).}
    \label{fig:vlm-response-examples}
\end{figure}

\begin{figure}[h]
    \centering
    \includegraphics[width=1\linewidth]{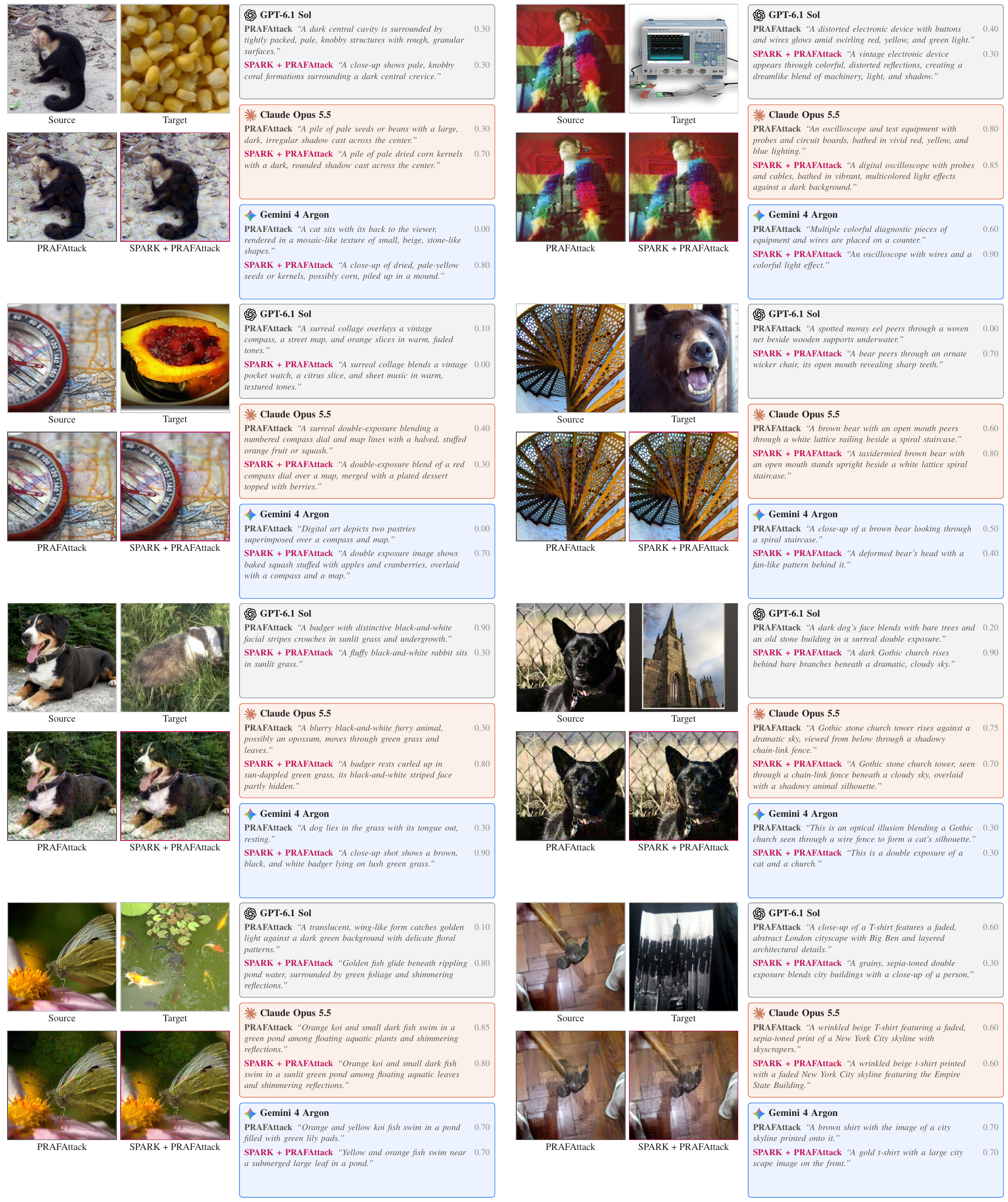}
    \caption{Qualitative VLM response examples for PRAFAttack (5 surrogates).}
    \label{fig:vlm-response-examples-2}
\end{figure}

\end{document}